\documentclass[a4paper,fleqn]{cas-dc}

\usepackage[authoryear]{natbib}

\usepackage{graphicx}%
\usepackage{multirow}%
\usepackage{amsmath,amssymb,amsfonts}%
\usepackage{amsthm}%
\usepackage[title]{appendix}%
\usepackage{textcomp}%
\usepackage{manyfoot}%
\usepackage{booktabs}%
\usepackage{algorithm}%
\usepackage{algorithmicx}%
\usepackage{algpseudocode}%
\usepackage{listings}%
\usepackage{xcolor}%
\usepackage{array}%
\usepackage{diagbox}%
\usepackage{hyperref}%
\usepackage[protrusion=true,expansion=true]{microtype}
\hypersetup{
  pdftitle={SPLIT: Separating Physical-Contact via Latent Arithmetic in Image-Based Tactile Sensors},
  pdfauthor={Wadhah Zai El Amri and Nicol\'as Navarro-Guerrero},
  pdfsubject={Robotics (cs.RO); Artificial Intelligence (cs.AI), Computer Vision and Pattern Recognition (cs.CV)},%
  pdfkeywords={Tactile Sensing, Real2Sim, Robotic Perception, Image-Based Tactile Sensors},%
  colorlinks=true,
  linktoc=all,
  linkcolor=black,
  citecolor=black,
  urlcolor=black,
}

\def\tsc#1{\csdef{#1}{\textsc{\lowercase{#1}}\xspace}}
\tsc{WGM}
\tsc{QE}

\newcolumntype{Y}[1]{>{\centering\arraybackslash}p{#1}}

\begin{document}
\let\WriteBookmarks\relax
\def\floatpagepagefraction{1}
\def\textpagefraction{.001}

\shorttitle{SPLIT: Separating Physical-Contact via Latent Arithmetic in Image-Based Tactile Sensors}

\shortauthors{Zai El Amri and Navarro-Guerrero}

\title [mode = title]{SPLIT: Separating Physical-Contact via Latent Arithmetic in Image-Based Tactile Sensors}

\author[1]{{Wadhah} {Zai El Amri}}[orcid=0000-0002-0238-4437]
\cormark[1]
\ead{wadhah.zai@l3s.de}
\credit{Code, Methodology, Investigation, Formal analysis, Data curation, Visualization, Writing}

\author[1]{{Nicol\'as} {Navarro-Guerrero}}[orcid=0000-0003-1164-5579]
\ead{nicolas.navarro.guerrero@gmail.com}
\credit{Writing, Formal analysis, Conceptualization, Supervision, Project administration}

\cortext[1]{Corresponding author}



\affiliation[1]{organization={Leibniz Universität Hannover, L3S Research Center},
            addressline={Appelstra\ss e 4},
            city={Hannover},
            postcode={30167},
            state={Lower Saxony},
            country={Germany}}

\begin{abstract}
Training machine learning models for robotic tactile sensing requires vast amounts of data, yet obtaining realistic interaction data remains a challenge due to physical complexity and variability. Simulating tactile sensors is thus a crucial step in accelerating progress. This paper presents SPLIT, a novel method for simulating image-based tactile sensors, with a primary focus on the DIGIT sensor. Central to our approach is a latent space arithmetic strategy that explicitly disentangles contact geometry from sensor-specific optical properties. Unlike methods that require recalibration for every new unit, this disentanglement allows SPLIT to adapt to diverse DIGIT backgrounds and even transfer data to distinct sensors like the GelSight R1.5 without full model retraining. Beyond this adaptability, our approach achieves faster inference speeds than existing alternatives. Furthermore, we provide a calibrated finite element method (FEM) soft-body mesh simulation with variable resolution, offering a tunable trade-off between speed and fidelity. Additionally, our algorithm supports bidirectional simulation, allowing for both the generation of realistic images from deformation meshes and the reconstruction of meshes from tactile images. This versatility makes SPLIT a valuable tool for accelerating progress in robotic tactile sensing research.
\end{abstract}





\maketitle


\section{Introduction}

The field of robotics has witnessed significant advancements in tactile sensing, with various new sensors emerging that offer enhanced capabilities for modern robots~\citep{Navarro-Guerrero2023VisuoHaptic}. These sensors are being increasingly deployed in diverse robotic solutions, showcasing a growing focus on their development and application. However, training machine learning algorithms on these sensors requires substantial amounts of data, which can be time-consuming and energy-intensive to collect using real robots. Therefore, simulating tactile sensors has become crucial for accelerating progress in this area.

This paper's main objective is to present a novel framework, SPLIT, that enables accurate and generalizable simulation of image-based tactile sensors. By harnessing latent space arithmetic to extract and transfer the fundamental properties encoded in the model's feature space, we directly address key barriers in tactile sensing research.
We benchmark our method against state-of-the-art simulation algorithms, focusing primarily on the DIGIT sensor. We selected DIGIT as our main testbed because it is a low-cost, widely adopted, and commercially available solution, ensuring our results are directly relevant to the broader research community.

Our comparison includes a rigorous evaluation of image fidelity and an in-depth analysis of inference times to assess the computational efficiency of each approach. Furthermore, we demonstrate the versatility of our algorithm by extending it to the GelSight R1.5, highlighting SPLIT's ability to generalize to other commercially available sensors with distinct optical features.
In addition to simulating images, we also showcase our method's ability to generate accurate meshes of the DIGIT sensor's deformation when interacting with various objects. This capability emphasizes our approach's potential to provide comprehensive and detailed bidirectional simulation of tactile interactions.
Finally, we validate SPLIT in a real-world scenario by predicting force values from tactile images, demonstrating its practical utility in robotic grasping applications.

In summary, the main contributions of this work are:
\begin{itemize}
    \item A simulation framework (SPLIT) that utilizes $\beta$-VAEs and latent vector arithmetic to explicitly disentangle contact geometry from sensor-specific optical properties.
    \item An efficient inference pipeline capable of generating high-fidelity tactile images from low-resolution meshes, achieving significant speedups over geometric baselines.
    \item A demonstration of cross-sensor generalization, enabling the synthesis of outputs for different unseen DIGIT units without manual calibration and even extending to distinct sensors like the GelSight R1.5 from a single deformation source.
    \item Practical validation of the learned representations through accurate 3D geometry reconstruction and robust force estimation on unseen sensors.
    \item The open-source release of a comprehensive real-world dataset, a calibrated soft-body simulation environment, and the full SPLIT codebase on our website:~\href{https://wzaielamri.github.io/publication/split}{\textcolor{gray}{wzaielamri.github.io/publication/split}}.
\end{itemize}


\section{Related Work}

Recent efforts have focused on developing algorithms that simulate image-based tactile sensors, such as DIGIT~\citep{Lambeta2020DIGIT} and GelSight R1.5~\footnote{\label{2022GelSight}GelSight R1.5: \url{https://www.gelsight.com} - Last accessed: 08.04.2026.}. For example, Taxim~\citep{Si2022Taxim} uses a second-order polynomial  function and example-based photometric stereo methods. However, this approach has limits, especially when generating DIGIT outputs. The imperfections in the DIGIT gel surface can cause inconsistent shadows.
To address these limitations, alternative approaches have been proposed. One such approach is FOTS~\citep{Zhao2024FOTS}, which uses multi-layer perceptrons (MLPs) for mapping deformation and planar methods to generate shadows. While this method claims to be faster and more accurate than Taxim, it requires thorough calibration for each new sensor. Moreover, it can be impractical when working with multiple sensors that exhibit subtle variations on a sub-millimeter level due to manufacturing.

Another key challenge in tactile sensing is the scalability and sensor interoperability, which is currently limited by the diversity of sensor outputs, morphologies, and data representations, making it challenging to use data from different sensor sources interchangeably. Several researchers have attempted to address this issue by developing frameworks for transferring knowledge between sensors. For example, Zai El Amri et al.~\citep{ZaiElAmri2025ACROSS} proposed a framework that enables the transfer of data from a signal-based tactile sensor (BioTac) to a vision-based tactile sensor (DIGIT). Their algorithm uses disentangled variational autoencoders ($\beta$-VAEs) to convert BioTac signals into corresponding mesh deformations, which are then used to generate DIGIT deformations and render output images using Taxim~\citep{Si2022Taxim}. However, this method relies heavily on the Taxim~\citep{Si2022Taxim} algorithm, requiring recalibration and fine-tuning for every sensor. Additionally, it cannot transfer DIGIT images to DIGIT meshes; thus, it is a unidirectional method.

In contrast to previous approaches discussed above that face calibration and bidirectionality challenges, such as \citep{ZaiElAmri2025ACROSS}, we provide a calibration-free optical simulator for image-based tactile sensors that outperforms state-of-the-art methods (i.e., Taxim~\citep{Si2022Taxim} and FOTS~\citep{Zhao2024FOTS}). Leveraging latent-space arithmetic and a significantly larger open dataset, our algorithm generalizes well both within the same sensor and across different sensors (e.g., DIGIT $\rightarrow$ GelSight). Furthermore, our algorithm is inherently flexible, allowing for easy extension to convert images into 3D meshes when needed.


\section{Methods}
\label{sec:methods}
\subsection{Overview}
To address the challenge of simulating image-based sensors like the DIGIT and GelSight R1.5, we developed a methodology that leverages disentangled latent representations to bridge the gap between surface deformation and optical response. Our approach begins with the acquisition of a real-world interaction dataset, which facilitates the empirical determination of the elastomer’s material properties. By encoding these physical characteristics into our modeling process, we enable the generation of accurate, physically consistent tactile images.
\begin{figure*}[!tbp]
  \centering
  \includegraphics[width=0.7\textwidth, trim=0cm 90mm 0cm 0cm, clip]{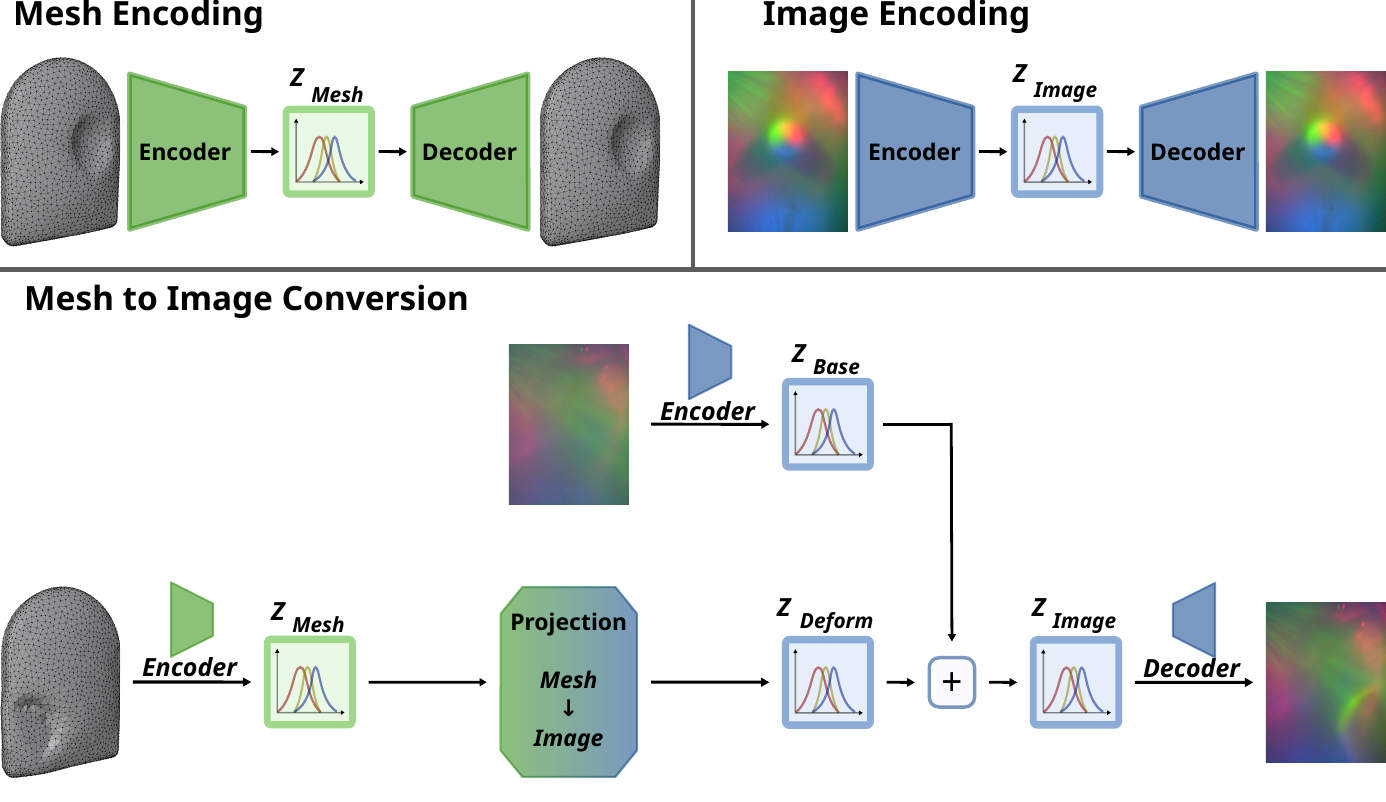}
  \caption{Schematic representation of the SPLIT framework. The pipeline proceeds in two stages: First, we train separate $\beta$-VAEs to learn compact, structured representations of meshes and images. Subsequently, we train a cross-modal projection network to map between these latent spaces. Notably, we employ latent space arithmetic to disentangle geometry from optics: During training, we subtract the reference background vector to force the network to learn pure deformation patterns. During inference, as illustrated, we add the latent vector of the target background to this deformation representation, enabling flexible, sensor-specific image generation. Note: The DIGIT camera is positioned behind the gel, resulting in a mirrored view relative to the mesh.}
  \label{fig:full_pipeline_networks}
\end{figure*}

The proposed pipeline, illustrated in Figure~\ref{fig:full_pipeline_networks}, utilizes a two-stage process for cross-modal encoding and decoding between meshes and images. First, we independently train two disentangled Variational Autoencoders ($\beta$-VAEs), one for meshes and one for images, selected for their continuous latent spaces to facilitate domain transfer~\citep{Narang2021SimtoReal, ZaiElAmri2025ACROSS}.

In the second stage, we link these modalities using a multi-layer perceptron (MLP) projection network trained to map the latent representation of 3D meshes ($Z_{Mesh}$) to a tactile deformation space ($Z_{Deform}$). Central to SPLIT is a latent space arithmetic strategy that disentangles geometry from optics. To define the training target for the network, we subtract the reference background vector ($Z_{Base}$), representing an undeformed sensor, from the ground-truth image latent vector ($Z_{Image}$). This forces the MLP to learn to predict only the pure deformation patterns ($Z_{Deform}$), invariant to sensor-specific optical characteristics. During inference, we reconstruct the final tactile image by simply adding a chosen background vector ($Z_{Base}$) to the predicted deformation vector. This additive reconstruction grants high adaptability: by swapping the $Z_{Base}$ vector during this step, we can simulate varying DIGIT backgrounds or transfer to distinct sensors with similar mesh geometry, such as the GelSight R1.5, without retraining. Crucially, this ability to explicitly exchange optical properties serves as a powerful mechanism for data augmentation and in downstream tasks.

\begin{figure*}[!tbp]
  \centering
  \includegraphics[width=0.7\textwidth, trim=0cm 0cm 0cm 140mm, clip]{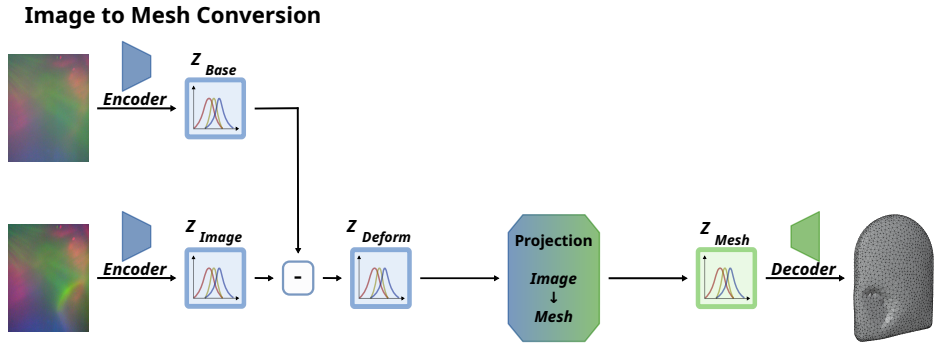}
  \caption{Schematic representation of our SPLIT method used for generating deformation meshes from DIGIT images.}
  \label{fig:image_mesh_networks}
\end{figure*}

We further extend the pipeline to support bidirectional inference, enabling the recovery of full 3D geometry from tactile images (Figure~\ref{fig:image_mesh_networks}). Mirroring our disentanglement strategy, we first isolate the deformation information by subtracting the background vector ($Z_{Base}$) from the input image latent vector ($Z_{Image}$). This resulting vector is then mapped to the mesh latent representation ($Z_{Mesh}$) via a reverse projection network.
Finally, we validate the practical utility of SPLIT in a downstream force estimation task.

The remainder of this section details the experimental setup, starting with the data acquisition workflow and physics calibration, followed by details regarding the specific experiments used to validate the framework’s performance. Specific architectural details, including layer configurations and training hyperparameters, are provided in the supplementary material on our website.


\subsection{Data Collection}
\label{sec:dataset}
To gather the necessary data for training and evaluating our simulation method, we collected an extensive dataset of sensor interactions. This dataset was created by recording the output of the DIGIT sensor, which was mounted on a UR5e robotic arm, as it interacted with various indenters of varying shapes and sizes. The DIGIT sensor incrementally moved by 0.1 mm between each frame capture. The data collection process involved systematically varying the DIGIT sensor's angle, force, and orientation in each new trajectory, capturing detailed interactions. To ensure the generalizability of our dataset, we utilized five different DIGIT sensors to account for variations in their gel and lighting properties.

\begin{figure}[!htb]
  \centering
  \includegraphics[width=1\columnwidth]{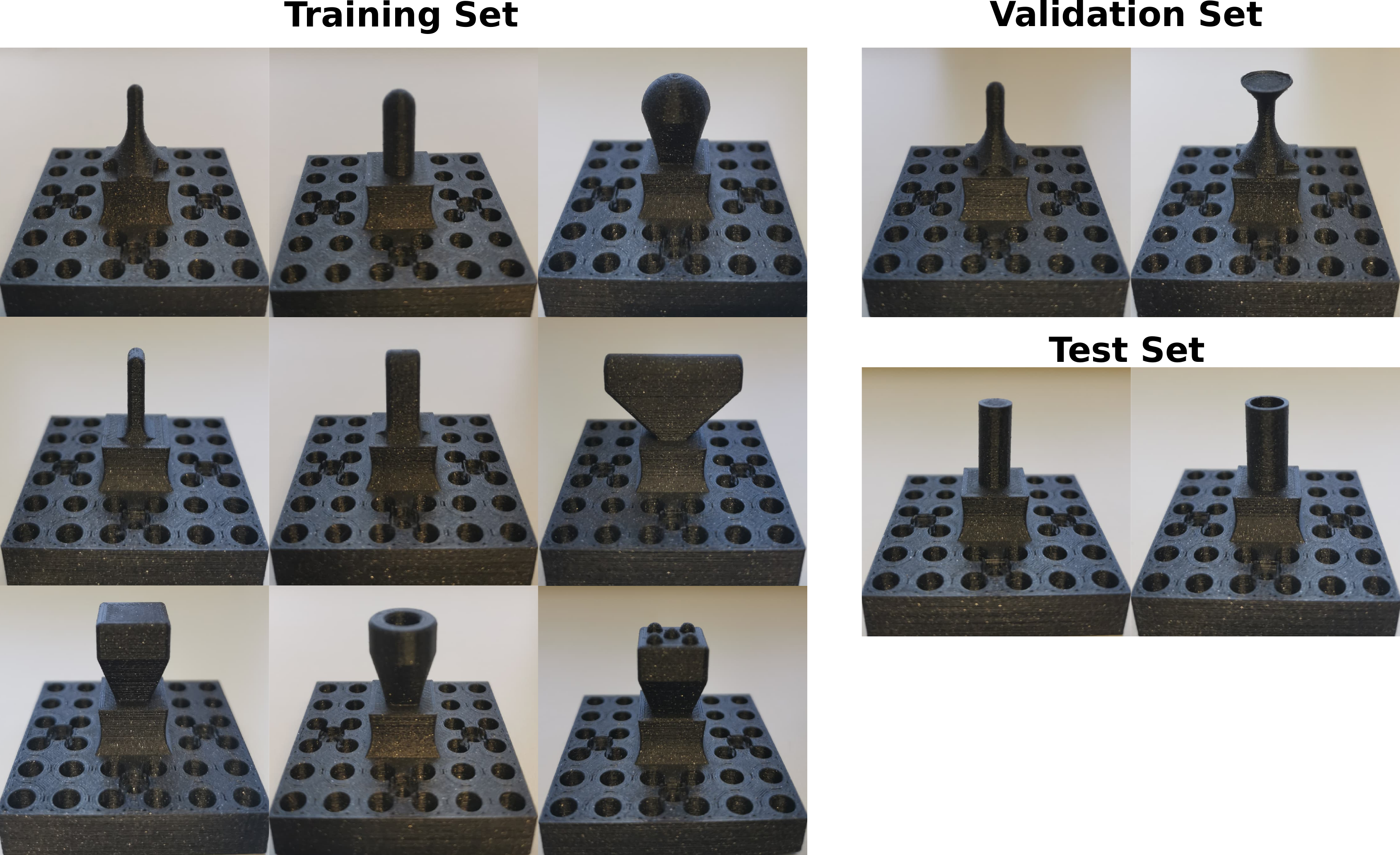}
    \caption{Visualization of the 13 distinct indenters used in the dataset collection.}
\label{fig:all_indenters}
\end{figure}

As a result, our dataset comprises over 65,000 unique trajectories of 13 distinct indenters, visualized in Figure~\ref{fig:all_indenters}, as well as more than 1 million contact images.
Our open-sourced dataset includes accurate force and torque values. These values were collected using a Robotiq FT 300-S force/torque sensor~\footnote{\label{force} Robotiq FT 300-S force/torque sensor: \url{https://robotiq.com/products/ft-300-force-torque-sensor} - Last accessed: 08.04.2026.}. This data encompasses full 6-axis measurements, providing detailed resolution on both shear ($F_x, F_y$) and normal ($F_z$) components along with torque vectors.

This rich dataset forms the foundation for training and evaluating our simulation method. Its large size promotes generalization in our pipeline and supports further research in related tasks. It also allows the reuse of our pipeline as pre-trained models without fine-tuning.
For our subsequent analysis, we selected a representative subset of 250,884 unique images. These images were selected in instances where the sensor was in contact with the object and subjected to forces ranging from 1 N to 13 N. We established this lower bound of 1 N because forces below this threshold yielded minimal or no visible deformation in the captured images, while the upper bound was set at 13 N to avoid damaging the gel with excessive force.

For all our experiments, we adopt a rigorous data-partitioning strategy to comprehensively evaluate the effectiveness and robustness of our proposed simulation method. Specifically, we employ a non-contiguous data-splitting, where the data is first segregated by trajectory, then further stratified by sensors and indenters. This separation ensures that the training, validation, and test sets are diverse. This strategy also reduces correlation among the sets. The dataset is divided into 189,433 training samples, 23,763 validation samples, 29,536 test samples in the first partition, and 8,152 test samples in the second partition.

To ensure our pipeline generalizes across different data distributions and to further augment the capacity of our image $\beta$-VAE, we supplemented our dataset with external open-source collections. We incorporated 82,463 samples from Fu et al.~\citep{Fu2024Touch} and 4,000 from Zhao et al.~\citep{Zhao2024Transferable}, bringing the total dataset size to 266,110 training and 33,549 validation samples. Furthermore, we employ a common data augmentation strategy, including rotation, flipping, adding noise, and varying color contrast and brightness. These augmentations enable our model to learn more robust features. As a result, the model can generalize to new, unseen data and eliminate the need for calibration typically required by state-of-the-art approaches.


\subsection{Physics Calibration} \label{sec:physics_calibration}
To train the mesh-encoding networks, our pipeline requires ground truth 3D representations of the gel's deformation that correspond to the collected real-world images. Since capturing the precise 3D surface topology of the elastomer during real-time contact is physically infeasible, we rely on physics-based simulation to reconstruct these geometries. By replicating the recorded real-world trajectories within a simulation environment, we can generate the exact deformation meshes resulting from each interaction.

For this implementation, we utilize the Isaac Gym simulation environment~\citep{Makoviychuk2021Isaac} to simulate the collected trajectories, leveraging its GPU acceleration capabilities. Although any soft-body capable simulator (e.g., Isaac Sim~\citep{mittal2023orbit} or MuJoCo~\citep{todorov2012mujoco}) can be employed.

The simulation proposed by Zai El Amri et al.~\citep{ZaiElAmri2025ACROSS} relies solely on theoretical values for the Poisson ratio and elasticity modulus of the DIGIT gel material, which may be prone to inaccuracies and discrepancies due to the complexities of real-world materials and manufacturing processes. In contrast, we explicitly mitigate these deviations by empirically determining these parameters, thereby ensuring a higher degree of physical fidelity in our simulation.

According to the manufacturer, the DIGIT gel has a durometer of 75 on the Shore 00 scale, equivalent to a range of $24$-$25$ on the Shore A scale. Using the following Gent's~\citep{Gent1958Relation} equation, we can approximate the Young's modulus based on the durometer value in Shore A scale:
\begin{equation}\label{eq:gent}
    E = \frac{0.0981 \times (56 + 7.62336\times S)}{0.137505 \times (254 - 2.54\times S)},
\end{equation}
where $S$ is the durometer value in Shore A scale and $E$ is the Young's modulus in megapascals ($MPa$). Applying Equation~\ref{eq:gent}, we determine the range of the elasticity modulus, ranging between 883.140 $kPa$ and 923.464 $kPa$.

In fact, a range of parameters can achieve alignment between deformation and the rendered image for the same elastomer. To further refine this alignment and validate our simulation, we employ SMAC~\citep{Lindauer2022SMAC3} by additionally including force readings ($F_x$, $F_y$, and $F_z$), which provide another metric to ensure precise alignment.
The simulation utilizes a Finite Element Method (FEM) approach, natively supported by Isaac Gym's soft-body solver, to compute the contact dynamics. Specifically, the environment calculates the nodal forces across the soft-body mesh by resolving the stress tensors derived from our empirically calibrated elasticity modulus, Poisson ratio, and friction coefficient.
We define a search space with the following bounds: an elasticity modulus between $838$ $kPa$ and $970$ $kPa$ (±$5\%$ of the experimental value calculated using Gent's Equation~\ref{eq:gent}), Poisson ratio ranging from $0.45$ to $0.5$, and the friction coefficient ranging from $0.5$ to $0.99$, both of which are common ranges for rubber-like materials like the DIGIT gel~\citep{Mott2009Limits,Persson2001Theory}.

We use the Hyperparameter Optimization Facade with Random Forest to minimize the following loss function:
\begin{equation}\label{eq:smac_loss}
    L = \frac{1}{N} \sum_{n=1}^{N} \frac{1}{T} \sum_{t=1}^{T} \left| F_{\text{real},t}^n - F_{\text{sim},t}^n \right|,
\end{equation}
where $F_{\text{real}}$ represents the norm of the real force values measured using the Robotiq FT-300S force/torque sensor~\textsuperscript{\ref{force}}, $F_{\text{sim}}$ denotes the force computed through the finite element method (FEM) in the simulation, $T$ is the length of the trajectory, and $N$ is the number of trajectories/environments in each trial. We use $25$ randomly selected trajectories in each trial.
After 4500 optimization trials, we obtain the following values for the elasticity modulus, Poisson ratio, and friction coefficient: $841.509$ $kPa$, $0.464$, and $0.987$, respectively. Using these parameters, we simulate the real sensor trajectories we collected. Our open-source simulation offers two different deformation qualities for the DIGIT gel: low-resolution with $6,103$ vertices and high-resolution with $80,744$ vertices. For all our subsequent experiments, we use the low-resolution meshes, as this already outperformed the baselines. This choice also allows us to evaluate our method's performance in a scenario with limited computational resources.


\subsection{Experiments}

\subsubsection{Simulating DIGIT Sensor Images}
To validate the performance of our novel SPLIT framework (Figure~\ref{fig:full_pipeline_networks}), we conducted two main comparative experiments. First, we compare its simulation capabilities against two state-of-the-art image-based simulation methods: Taxim~\citep{Si2022Taxim} and FOTS~\citep{Zhao2024FOTS}.
Second, to isolate the contributions of specific architectural choices, we evaluate two internal variations:
\begin{itemize}
    \item \textbf{Architecture Validation:} We compare our $\beta$-VAE-based pipeline against a traditional end-to-end autoencoder trained to generate images directly from 3D DIGIT mesh vertices, demonstrating the benefits of our structured latent space.
    \item \textbf{Disentanglement Validation:} We verify the impact of our latent arithmetic strategy by testing SPLIT under two conditions. We compare a direct mapping approach (denoted \textit{NOSPLIT}), which maps the mesh representation directly to the full image latent ($Z_{Image}$), against our proposed arithmetic latent mapping strategy (denoted \textit{SPLIT}), which predicts a deformation vector ($Z_{Deform}$) and explicitly adds a target background ($Z_{Base}$).
\end{itemize}

To quantitatively evaluate the generation quality across these experiments, we report the Mean Absolute Error ($\ell_1$), Mean Squared Error (MSE), Structural Similarity Index Measure (SSIM), and Peak Signal-to-Noise Ratio (PSNR).

Regarding the external baselines, Taxim and FOTS typically calculate a height map from the 3D mesh vertices of the touched object. This height map is then used as input. However, in some scenarios, we may lack prior knowledge of the object's structure, which is often arbitrary and unknown in real-world environments, and instead, only have information about the sensor's deformation. Therefore, we compute the height map using Pyrender~\footnote{\label{Pyrender}Pyrender: Easy-to-Use glTF 2.0-Compliant OpenGL Renderer for Visualization of 3D Scenes: \url{https://github.com/mmatl/pyrender} - Last accessed: 08.04.2026.} from the deformation mesh of the DIGIT sensor, a more general and efficient approach that yields results similar to those obtained by calculating the height map from the shape of the touched 3D object. When rendering Taxim and FOTS images, we employ high-resolution DIGIT meshes with $80,744$ vertices.
In contrast, our framework is designed to work with low-resolution DIGIT meshes, which have only $6,103$ vertices. Crucially, the meshes for both our method and the baselines originate from the same simulation environment; this ensures a consistent physical basis for comparison while demonstrating the robustness of our neural networks in extracting useful information from low-resolution representations.

Both baselines, i.e., Taxim and FOTS, generate images that can be misaligned on the $x$ and $y$ axes. This results from their inability to account for minor, sub-millimeter changes in the DIGIT manufacturing gel form and internal camera position, variations common to such a sensor. Zhao et al.~\citep{Zhao2024FOTS} also acknowledge this limitation: \textit{``Due to the accuracy of the operation with the real sensor, the ground truth tactile images do not match perfectly with the simulated images, promoting us to manually align the images using GIMP''}. Although we carefully calibrated the Pyrender camera position to align simulated and real images for 5 DIGIT sensors across all 13 indenters and 1000 trajectories each, achieving perfect alignment in every case proved challenging. Furthermore, we only update the background sensor image and use the authors' \citep{Si2022Taxim,Zhao2024FOTS} default parameters and calibration files.


\subsubsection{Intra-Sensor Generalization: Disentanglement and Target Alignment}\label{sec:methods_intra_sensor}

To comprehensively evaluate the disentanglement mechanism within SPLIT, we assess the image $\beta$-VAE and our latent arithmetic strategy through a cross-sensor optical style transfer task. Unlike the full pipeline which bridges modalities, this experiment operates within the visual domain, manipulating only pixel-derived latent representations. This validates whether the model can successfully isolate deformation geometry ($Z_{Deform}$) from sensor-specific optical characteristics ($Z_{Base}$) and generalize to novel domains. We adopt a two-stage strategy, addressing both quantitative statistical alignment and qualitative generalization to complex geometries.

\textbf{Quantitative Validation (Controlled Source $\to$ Unseen Target):} First, we quantified the method's capacity to transfer to the optical style of a novel, previously unseen DIGIT sensor while preserving the contact geometry of a known source. We utilized a subset of our internal dataset (Sensors 0-4) as the \textit{Source Domain} and two subsets of the open-source YCB-Slide dataset~\citep{Suresh2023MidasTouch} as the \textit{Target Domain} (Sensors A-B). This configuration allows for robust statistical evaluation: the source dataset contains controlled, known indentations, while the YCB-Slide dataset provides the ground-truth reference for the optical distribution of the target sensors.

To conduct this evaluation, we randomly sampled 1,000 recordings from each of the five source sensors (totaling 5,000 source images) and 1,000 recordings from each of the two target sensors (totaling 2,000 target images from the YCB-Slide dataset). We performed latent space disentanglement by first encoding the source images to isolate their deformation vectors ($Z_{Deform}^{Source}$). Specifically, for every image originating from a specific source sensor $i$, we subtracted the corresponding background reference vector ($Z_{Base}^{Source,i}$) from the full image latent ($Z_{Image}^{Source,i}$):
\begin{equation}
Z_{Deform}^{Source} = Z_{Image}^{Source,i} - Z_{Base}^{Source,i}
\end{equation}
Subsequently, we projected these deformations into the target domain by adding the encoded background representations ($Z_{Base}^{Target,j}$) of a target sensor unit $j$:
\begin{equation}
Z_{Synthetic}^{i\rightarrow j} = Z_{Deform}^{Source} + Z_{Base}^{Target,j}
\end{equation}
As no paired ground truth exists for this specific synthetic combination (source geometry with target optics), we evaluated generation fidelity using distribution-level metrics.
We utilized Histogram Intersection to measure global color fidelity. This metric is bounded in $[0, 1]$, where higher values indicate superior distributional overlap. To quantify perceptual texture similarity, we computed the VGG-19 Style Loss (Gram Matrix distance~\citep{Gatys2016Image}) between the synthetic output and real target data. While this metric is lower-bounded by 0 (perfect alignment), it is unbounded above. Therefore, we interpret it as a relative metric, where lower values (compared to the baseline) indicate a reduction in the perceptual domain gap.
Additionally, we analyzed the manifold alignment of these synthetic latent vectors using t-Distributed Stochastic Neighbor Embedding (t-SNE). To ensure visual clarity, the t-SNE analysis was restricted to a representative subset of two source sensors alongside the two target sensors, verifying if the synthetic embeddings successfully migrated to the target domain's cluster.

\textbf{Qualitative Generalization (Unseen Source Sensor \& Geometry $\to$ Unseen Target Sensor):} In this second stage, we conducted a qualitative test focusing on the transfer of unknown indentations to unknown sensors. We utilized one subset of the YCB-Slide dataset as the source domain with an unseen DIGIT, introducing complex, irregular contact geometries absent from our training distribution, and projected these unseen interactions onto the optical background of another distinct, unseen DIGIT sensor.
This configuration represents a challenging scenario: synthesizing tactile images where neither the contact shape (content) nor the sensor characteristics (style) were seen during training. As no ground truth exists for this specific combination of novel geometry and novel optics, we relied on qualitative visual inspection to verify that the network successfully preserves the intricate structural details of touches while accurately rendering the image with the unique lighting gradients and manufacturing artifacts of the target hardware.


\subsubsection{Cross-Sensor Generalization: Adapting to GelSight R1.5}
\label{sec:methods_cross_sensor}

We challenge the SPLIT framework with a significant domain shift by incorporating the GelSight R1.5 into our analysis. This experiment utilizes the complete mesh-to-image pipeline to conduct a rigorous test of cross-hardware generalization. The GelSight R1.5 possesses drastically different optical properties, resolution, and color profiles than the DIGIT, requiring the system to synthesize images for a completely new sensor using only the deformation geometry derived from DIGIT meshes.

To address this, we employ a transfer learning strategy that leverages the architectural modularity of SPLIT. We fine-tune the pre-trained Image $\beta$-VAE using the \textit{ObjectFolder-Real} dataset~\citep{Gao2023Objectfolder}, which comprises real-world recordings from the GelSight R1.5. The objective of this targeted fine-tuning is to adapt the decoder to the visual domain of the new sensor, while attempting to align its latent space with the structural representations already learned from the DIGIT data.

Crucially, the rest of the pipeline, specifically the mesh $\beta$-VAE and the cross-modal projection MLP, remains unchanged. We rely on the disentanglement properties of our latent space arithmetic to bridge the gap: during inference, we project the DIGIT deformation latent vector ($Z_{Deform}$) and inject the latent representation of a reference GelSight R1.5 background ($Z_{Base}$) replacing the DIGIT background. This experimental design allows us to assess the feasibility of adapting the simulation to new hardware by modifying only the optical components, treating the learned geometric encoding as a sensor-agnostic foundation.

Through this setup, we aim to validate the framework's capacity to translate existing DIGIT datasets into the GelSight domain, quantitatively evaluating the potential to reduce dependencies on large-scale physical data collection.


\subsubsection{Computational Efficiency and Real-Time Feasibility}
To assess the suitability of SPLIT for real-time robotic control loops, we benchmark its inference latency against established simulation baselines. All performance metrics were recorded on a standardized workstation equipped with an AMD EPYC 7452 CPU and an NVIDIA A6000 GPU.

We measure the average wall-clock time in seconds (s) required to generate a single tactile image frame. Our comparison includes:
\begin{itemize}
    \item The direct Isaac Gym soft-body simulation, which inherently necessitates GPU acceleration.
    \item The rendering time for Taxim~\citep{Si2022Taxim} and FOTS~\citep{Zhao2024FOTS} (implemented via Pyrender~\textsuperscript{\ref{Pyrender}}), executed on the CPU.
    \item The rendering time for our proposed SPLIT framework, also running entirely on the CPU.
\end{itemize}

The objective of this configuration is to quantify the computational efficiency of each approach, benchmarking the run-time performance of our method against established baselines on a consistent hardware architecture.


\subsubsection{Reconstructing DIGIT Sensor Meshes}
\label{sec:Methods_Experiments_Reconstructing_DIGIT_Sensor_Meshes}
Another key application of the proposed method is reconstructing the full 3D meshes through latent space mapping via our projection network. This inference process follows the pipeline's bidirectional architecture: First, input tactile images are encoded into the image latent space ($Z_{Image}$). To isolate the geometry-specific features, we subtract the reference background vector ($Z_{Base}$) from $Z_{Image}$. The resulting deformation vector ($Z_{Deform}$) is then mapped to the mesh latent representation ($Z_{Mesh}$) via the reverse projection network and finally decoded to recover the 3D mesh. To evaluate generalization, we test on unseen DIGIT sensor images. We benchmark our performance against the method proposed by Zhu et al.~\citep{Zhu2022Learning}, which targets the GelSlim sensor using a similar image-to-mesh translation strategy but relies on a differential renderer and synthetic training pairs. In contrast, our method leverages a digital twin simulation to pair real images with physically simulated meshes, which serve as the ground-truth geometries for our evaluation. As detailed in Section~\ref{sec:physics_calibration}, these meshes are acquired by replicating the exact real-world sensor trajectories within our calibrated Isaac Gym soft-body simulation environment.

To isolate the impact of our disentanglement strategy, we compare two projection approaches:

\begin{itemize}
    \item \textbf{Direct Mapping} (denoted \textit{NOSPLIT}): Maps the raw, entangled image latent vector ($Z_{Image}$) directly to the mesh embedding ($Z_{Mesh}$), without explicit background separation.
    \item \textbf{Arithmetic Latent Mapping} (denoted \textit{SPLIT}): Employs latent arithmetic to subtract the background vector ($Z_{Base}$) from the image representation ($Z_{Image}$). This yields an isolated deformation vector ($Z_{Deform}$), which serves as the input for the reverse mesh projection network.
\end{itemize}

We measure the geometric deviation between the predicted and ground-truth meshes using the Root Mean Squared Error (RMSE) and Euclidean distance, reporting all values in millimeters (mm).

Additionally, to assess the physical consistency of the reconstructed meshes under natural contact conditions, we qualitatively evaluate the framework using real tactile images captured during physical interactions with various objects.
Furthermore, to investigate the stability of the bidirectional pipeline for potential long-term temporal state estimation tasks, we designed a continuous cyclic reconstruction experiment (Image $\rightarrow$ Mesh $\rightarrow$ Image). In this setup, the output of one modality is recursively fed back as the input to the other for 40 consecutive iterations. We quantify the manifestation of cumulative error and signal degradation across these cycles using two sets of metrics: absolute deviation metrics (measuring Image SSIM and Mesh RMSE relative to the initial real-world input) and cycle-to-cycle drift metrics (measuring the difference between consecutive cycles $N$ and $N-1$).


\subsubsection{Downstream Task: Force Estimation}

Accurate force estimation is fundamental to robust robotic manipulation, enabling feedback loops that allow agents to adjust grip strength, prevent slippage, and safely handle deformable or fragile objects. To demonstrate that SPLIT can be used to estimate shear and normal forces from tactile images, and to validate that it preserves the physical fidelity required for such tasks, we evaluate the system’s ability to predict 3-DoF vectors ($F_x, F_y, F_z$) from tactile inputs.

We compare seven distinct architectural configurations to isolate the benefits of our latent representations:
\begin{itemize}
    \item \textbf{End-to-End Baselines:} We train supervised networks to predict forces directly from high-dimensional raw data. We evaluate two baselines: \textbf{Raw Vertices} (denoted \textit{Raw Vertices}) and \textbf{Raw Images} (denoted \textit{Raw Images}). These networks utilize the encoders from our pre-trained $\beta$-VAEs, with the probabilistic bottleneck replaced by a dense output layer predicting ($F_x, F_y, F_z$).

    \item \textbf{Latent Space Embedding:} We train MLPs to regress force values directly from the fixed latent embeddings and assess their information content. We evaluate three configurations:
    \begin{itemize}
        \item \textbf{Mesh Latent Vector} (denoted \textit{Mesh Latent}): Using the embedding ($Z_{Mesh}$) from the 3D mesh $\beta$-VAE.
        \item \textbf{Direct Image Latent Vector} (denoted \textit{Image Latent} (\textit{NOSPLIT})): Using the full, entangled visual state ($Z_{Image}$) which includes optical background properties.
        \item \textbf{Arithmetic Image Latent Vector} (denoted \textit{Image Latent} (\textit{SPLIT})): Using the isolated deformation signal ($Z_{Deform}$) obtained via latent subtraction, testing if removing optical information improves force estimation.
    \end{itemize}

    \item \textbf{Cross-Modal Projection:} Finally, we evaluate our full pipeline by projecting image latents into the mesh latent space ($Z_{Mesh}$) prior to force prediction. We test two mapping strategies:
    \begin{itemize}
        \item \textbf{Direct Mapping} (denoted \textit{Projected Mesh} (\textit{NOSPLIT})): The raw image latent ($Z_{Image}$) is projected directly to the mesh embedding using a projection network trained without disentanglement (Section~\ref{sec:Methods_Experiments_Reconstructing_DIGIT_Sensor_Meshes}).
        \item \textbf{Arithmetic Latent Mapping} (denoted \textit{Projected Mesh} (\textit{SPLIT})): The image latent is first processed via latent arithmetic to isolate the deformation vector ($Z_{Deform}$), which is then projected to the mesh embedding ($Z_{Mesh}$) using our disentangled projection network (Section~\ref{sec:Methods_Experiments_Reconstructing_DIGIT_Sensor_Meshes}).
    \end{itemize}
\end{itemize}

The recorded forces span a range of $1$ to $13$ N for the normal component ($F_z$) and $-3$ to $3$ N for the lateral components ($F_x$ and $F_y$) as described in Section \ref{sec:dataset}. To ensure the reliability of our results and assess robustness to domain shifts, we evaluate performance on a test set of 8,152 unseen trajectories involving unseen indenters and sensors. All experiments are repeated with 20 distinct random seeds. We report the average MAE and its standard deviation in Newtons (N), calculated across all force components ($F_x$, $F_y$, and $F_z$) and the vector magnitudes $||F||_2$ of the predicted and ground-truth forces.


\section{Results and Discussion}
\label{sec:result}
\subsection{Simulating DIGIT Sensor Images}
\label{sec:DIGIT_Images}

To quantitatively assess the fidelity of the simulated DIGIT sensor images, we compare our approach against the baseline methods using standard image similarity metrics. The full experimental results are summarized in Table~\ref{tab:results_digit}, which includes the Mean Absolute Error ($\ell_1$), Mean Squared Error (MSE), Structural Similarity Index Measure (SSIM), and Peak Signal-to-Noise Ratio (PSNR).

\begin{table*}[tbp]
\caption{Quantitative comparison of the proposed SPLIT and NOSPLIT architectures against state-of-the-art baselines across two levels of generalization: unseen trajectories (top) and unseen hardware units (bottom).}
\label{tab:results_digit}
\centering
\renewcommand{\arraystretch}{1.0}
{
\begin{tabular}{@{}p{5cm}|Y{22mm}|Y{22mm}|Y{22mm}|Y{22mm}@{}}
\toprule
\multirow{2}{*}{\textbf{Methods}} & \textbf{$\boldsymbol{\ell_1}$ $\downarrow$} &  \textbf{MSE $\downarrow$} & \textbf{SSIM $\uparrow$} & \textbf{PSNR $\uparrow$}  \\  \cmidrule{2-5}
& \multicolumn{4}{Y{100mm}@{}}{Tested on unseen trajectories: 29536 samples} \\ \midrule \hline
\textbf{End-to-end Autoencoder} &  22.129  &  \phantom{0}864.851  &  0.769  & 18.855 \\ \hline
\textbf{Taxim~\citep{Si2022Taxim}} & 20.380 & \phantom{0}852.943 & 0.745 & 19.074 \\ \hline
\textbf{FOTS~\citep{Zhao2024FOTS}} &  30.110 & 1340.051 & 0.698 & 16.960  \\ \hline
\textbf{NOSPLIT} & 15.758  & \phantom{0}539.956 &  0.817 & 21.744 \\ \hline
\textbf{SPLIT} &  \textbf{\phantom{0}5.911} &  \textbf{\phantom{0}82.672} &  \textbf{0.872} & \textbf{29.801} \\ \hline
\midrule
& \multicolumn{4}{Y{100mm}@{}}{Tested on unseen trajectories, sensor and indenters: 8152 samples} \\ \midrule \hline
\textbf{End-to-end Autoencoder} &  22.770  &  \phantom{0}938.818  &  0.787  &  18.433 \\ \hline
\textbf{Taxim~\citep{Si2022Taxim}} &  19.628  &  \phantom{0}792.485  & 0.737  & 19.338 \\ \hline
\textbf{FOTS~\citep{Zhao2024FOTS}} & 29.487 & 1262.684 & 0.693 & 17.174  \\ \hline
\textbf{NOSPLIT} &  22.621 &  \phantom{0}947.700 & 0.804  & 18.396 \\ \hline
\textbf{SPLIT} &  \textbf{\phantom{0}8.584}&  \textbf{\phantom{0}159.582} &  \textbf{0.845} &  \textbf{26.581}
\\ \hline
\end{tabular}
}
\end{table*}
\begin{figure}[!htb]
  \centering
    \includegraphics[width=1\columnwidth]{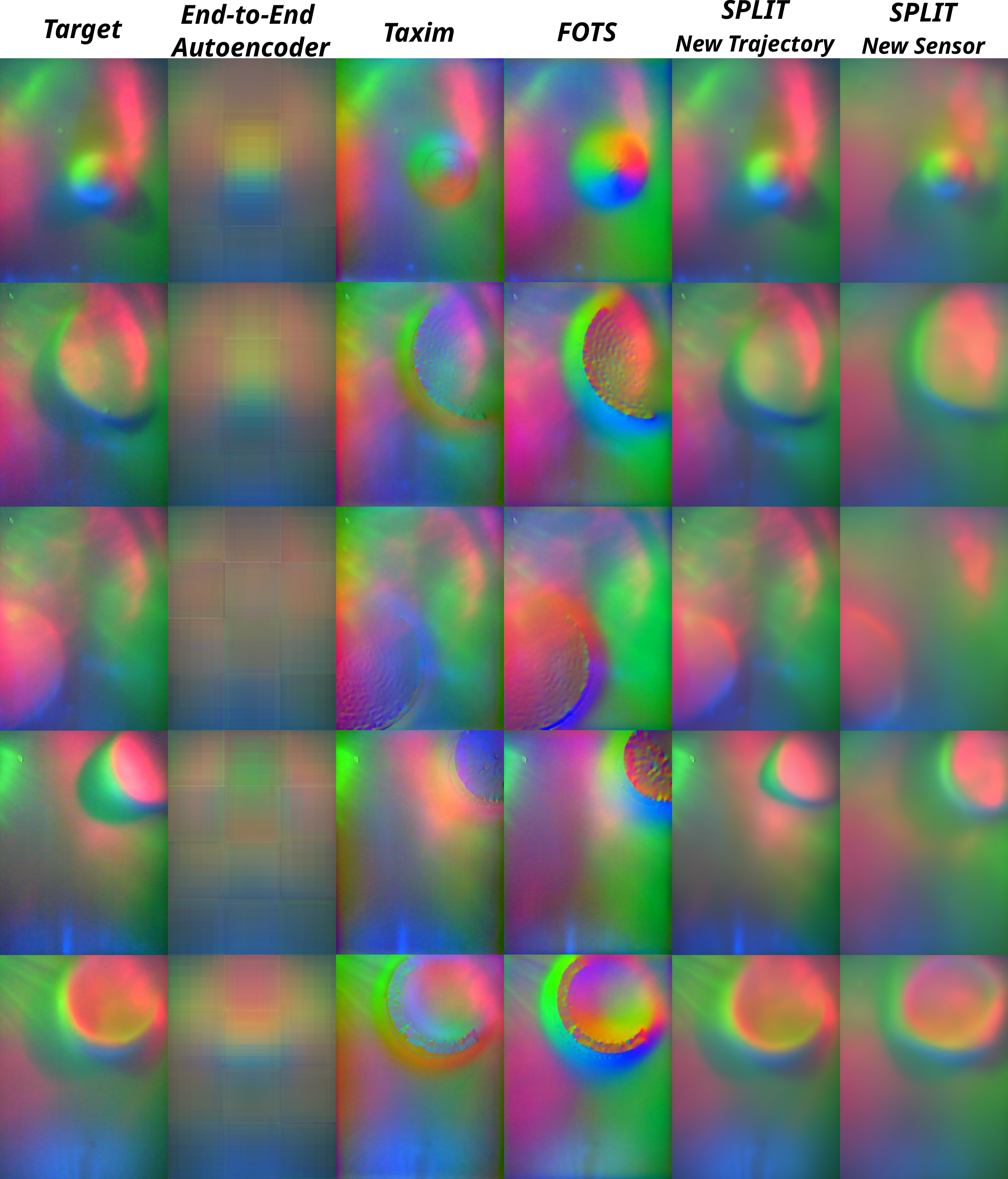}
    \caption{Comparison of predicted DIGIT images using different methods. The columns show: (1) real DIGIT images, (2) predicted images using an end-to-end autoencoder network, (3) predicted images using Taxim~\citep{Si2022Taxim}, (4) predicted images using FOTS~\citep{Zhao2024FOTS}, (5) predicted images using SPLIT (Arithmetic), and (6) predicted images using SPLIT, projected onto the background of an unseen DIGIT sensor.}
    \label{fig:results_digit}
\end{figure}

Table~\ref{tab:results_digit} and Figure~\ref{fig:results_digit} present the quantitative and qualitative performance of our framework. The proposed method generates highly accurate outputs for unseen deformations and sensor backgrounds, outperforming existing simulation baselines. Since Taxim~\citep{Si2022Taxim} and FOTS~\citep{Zhao2024FOTS} simulations depend solely on geometric depth maps, artifacts, such as those visible in Figure~\ref{fig:results_digit}, consistently appear in their outputs, even though these methods attempt to mitigate them via Gaussian smoothing. In contrast, our approach demonstrates strong independence from these simulation limitations. By leveraging latent space vector arithmetic, SPLIT requires only a single background sensor image, i.e., an image of the undeformed sensor, to adapt to new hardware, thereby eliminating the need for the tedious calibration or manual alignment often required by other state-of-the-art-baselines. Moreover, SPLIT achieves these superior results while utilizing low-resolution meshes ($6,103$ vertices), whereas the baselines rely on high-resolution inputs ($80,744$ vertices), highlighting the efficiency of our neural approach.

Our results further validate the architectural contributions of the SPLIT framework. First, regarding the architecture, our $\beta$-VAE-based pipeline significantly outperforms the traditional end-to-end autoencoder baseline across all reported metrics. This confirms that learning a structured, continuous latent representation facilitates more robust domain transfer than a direct vertex-to-pixel regression. Second, the results demonstrate the critical advantage of our disentanglement strategy. The Arithmetic Latent Mapping (denoted SPLIT) yields superior generation quality compared to the Direct Mapping approach (denoted NOSPLIT). This proves that explicitly isolating the deformation signal ($Z_{Deform}$) and subsequently incorporating the background ($Z_{Base}$) via vector addition is more effective than forcing the network to learn the full visual state ($Z_{Image}$) directly from the mesh representation ($Z_{Mesh}$).

\subsection{Intra-Sensor Generalization: Disentanglement and Target Alignment}
\label{sec:results_intra_sensor}

To validate the efficacy of the SPLIT framework, we conducted a comprehensive analysis of the disentangled latent spaces and the visual fidelity of the generated tactile images. Our evaluation follows the two-stage strategy defined in Section~\ref{sec:methods_intra_sensor}, assessing both the statistical alignment of controlled data and the qualitative generalization to complex, unseen DIGIT backgrounds.

\textbf{Quantitative Validation (Controlled
Source $\to$ Unseen Target):}
We begin by quantifying the domain alignment capabilities of our method across all source sensors and target sensors. To measure this, we computed histogram intersection similarities and VGG-based style losses (Table~\ref{tab:quantitative_metrics}). To demonstrate the improvement, we evaluated two scenarios: the baseline discrepancy between the original, unmodified source and target datasets (labeled ``Real''), and the resulting alignment between our synthetically adapted images and the target data (labeled ``Synth.''). This comparison validates the reduction in domain shift achieved by our method.

\begin{table*}[!tbp]
\centering
\caption{Quantitative comparison of domain alignment. We evaluate the distribution gap between Source sensors (0–4) and Target sensors (A–B). ``Real'' denotes the baseline metrics calculated between the raw, unmodified Source and Target datasets. ``Synth.'' denotes the metrics calculated between the synthetically adapted Source images (projected to the Target style) and the real Target data. Higher Histogram Similarity and lower Style Distance indicate better alignment with the target domain.}
\label{tab:quantitative_metrics}
\centering
\renewcommand{\arraystretch}{1.2} 

\begin{tabular}{l|Y{12mm}|Y{12mm}|Y{12mm}|Y{12mm}|Y{12mm}|Y{12mm}|Y{12mm}|Y{12mm}}
\toprule
\textbf{Comparison} & \multicolumn{4}{c|}{\textbf{Hist. Correlation} ($\uparrow$)} & \multicolumn{4}{c}{\textbf{Style Distance} ($\downarrow$)} \\
\hline

\multirow{2}{*}{\diagbox[width=8em]{Source}{Target}}
& \multicolumn{2}{c|}{Sensor A} & \multicolumn{2}{c|}{Sensor B}
& \multicolumn{2}{c|}{Sensor A} & \multicolumn{2}{c}{Sensor B} \\
\cmidrule{2-9}

& \textit{Real} & \textit{Synth.} & \textit{Real} & \textit{Synth.}
& \textit{Real} & \textit{Synth.} & \textit{Real} & \textit{Synth.} \\
\hline

Sensor 0 & 0.828 & \textbf{0.949} & 0.776 & \textbf{0.915} & 0.119 & \textbf{0.066}  & 0.126 & \textbf{0.073} \\ \hline
Sensor 1 & 0.921 & \textbf{0.933} & 0.761 & \textbf{0.946} & \textbf{0.048} & 0.061  & \textbf{0.072} & 0.074 \\ \hline
Sensor 2 & 0.831 & \textbf{0.950} & 0.738 & \textbf{0.930} & 0.128 & \textbf{0.065}  & 0.112 & \textbf{0.081} \\ \hline
Sensor 3 & 0.774 & \textbf{0.945} & 0.693 & \textbf{0.934} & 0.104 & \textbf{0.064}  & 0.090 & \textbf{0.077} \\ \hline
Sensor 4 & 0.834 & \textbf{0.942} & 0.653 & \textbf{0.946} & 0.069 & \textbf{0.063}  & 0.102 & \textbf{0.074} \\
\hline \hline
Average & 0.838 & \textbf{0.944} & 0.724 & \textbf{0.934} & 0.094 & \textbf{0.064}  & 0.100 & \textbf{0.076} \\
\hline
\end{tabular}
\end{table*}

Our latent arithmetic strategy yields a substantial improvement in statistical alignment relative to the raw sensor baselines, ensuring the synthetic images share the same statistical properties (lighting and texture) as the real target: the average histogram correlation improved from an unaligned reference of $0.838$ to $0.944$ for target domain A and from $0.724$ to $0.934$ for target domain B, indicating that the global color distributions and lighting intensities of the synthetic images closely match the real sensors.
Additionally, the synthetic images achieved significantly lower average VGG Style Distances ($0.064$ for Target A and $0.076$ for Target B) compared to the baselines ($0.094$ and $0.100$). This corresponds to a relative error reduction of approximately 32\% and 24\%, respectively, confirming a measurable reduction in the perceptual domain gap. Notably, Sensor 1 exhibited significantly lower baseline Style Distances ($0.048$ for Target A and $0.072$ for Target B) than the other source units. This suggests an incidental statistical alignment in global parameters, such as brightness distributions and contrast levels, between Sensor 1 and the targets. We attribute this partly to the metric's reliance on Gram matrices derived from an ImageNet pretrained VGG-19 network. The domain gap between natural images and elastomer surfaces can occasionally cause distinct tactile textures to map to similar feature correlations.
Crucially, however, our method converged the synthetic output of all sensors, including Sensor 1, to a uniform performance level. By stabilizing the Style Distance to $\approx 0.064$ for A and $\approx 0.076$ for B, SPLIT acts as an effective domain normalizer. It aligns the distinct optical profiles of diverse sensor units, converting varying background textures and lighting conditions into a single, consistent target representation. This capability effectively bridges large domain gaps (as seen in Sensors 0, 2, 3, 4) while ensuring consistent simulation quality even for sources that require less adaptation.

\begin{figure}[!htb]
\centering
\includegraphics[width=\linewidth]{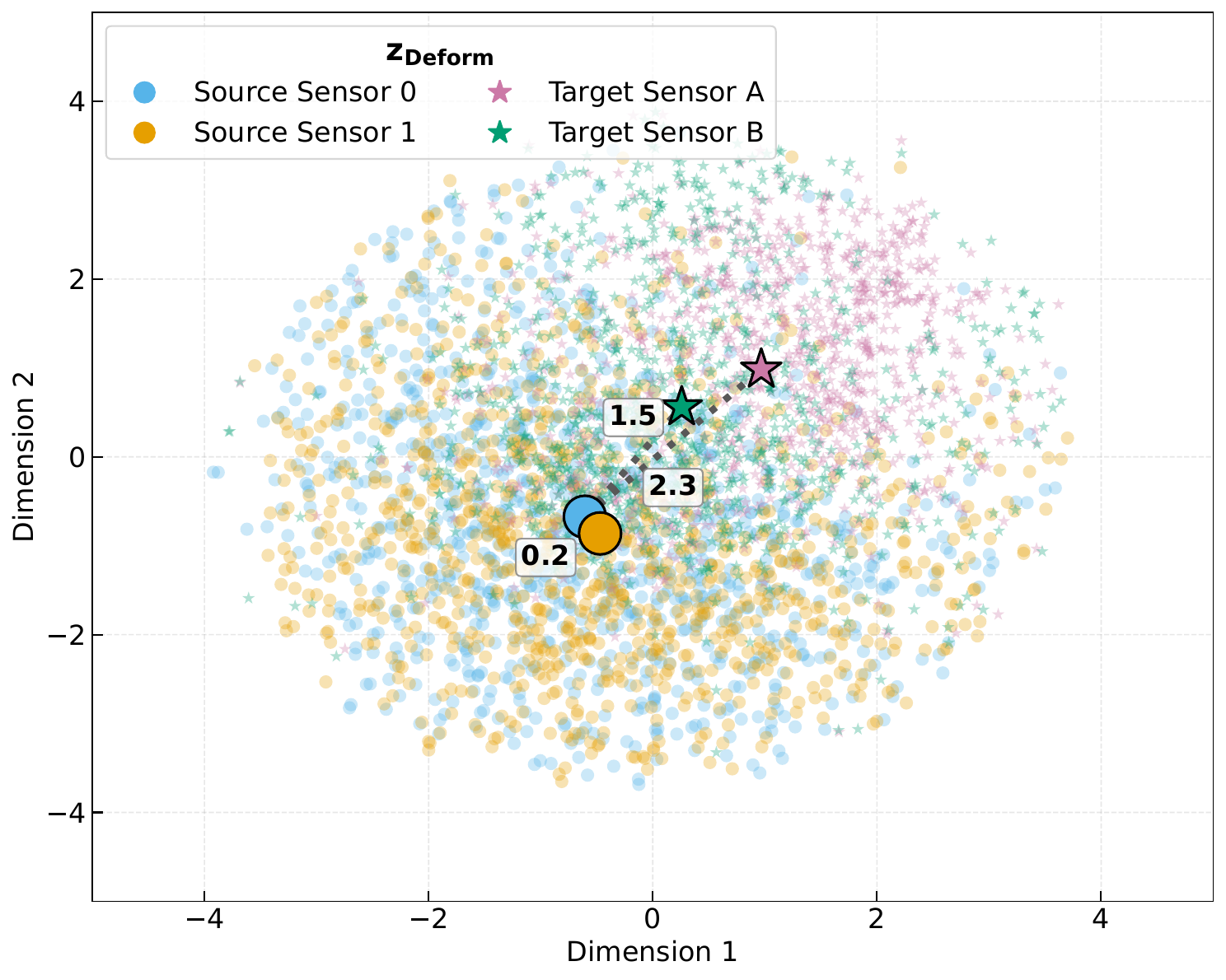}
\caption{Latent Space Distribution ($Z_{Deform}$): Geometry Invariance Check.}
\label{fig:tsne_z_deform}
\end{figure}

To better understand the structural basis of these improvements, we visualized the learned latent spaces using t-SNE for Source sensors 0-1 and Target sensors A-B. We first analyzed the deformation space ($Z_{Deform}$) to verify that the network encodes contact geometry independently of sensor hardware. Figure~\ref{fig:tsne_z_deform} visualizes the embedding of the deformation latent vectors for a representative subset of source and target sensors.

The plot reveals that the distributions for Sensor 0 and Sensor 1 (Source) overlap with the target sensors. We note that a slight distance exists between the target and source centroids, attributable to the fact that the target dataset consists of unseen indenters. Nevertheless, this structural alignment confirms that the learned representation remains independent of the sensor hardware itself and all $Z_{Deform}$ vectors effectively build a single cluster.

\begin{figure}[!htb]
\centering
\includegraphics[width=\linewidth]{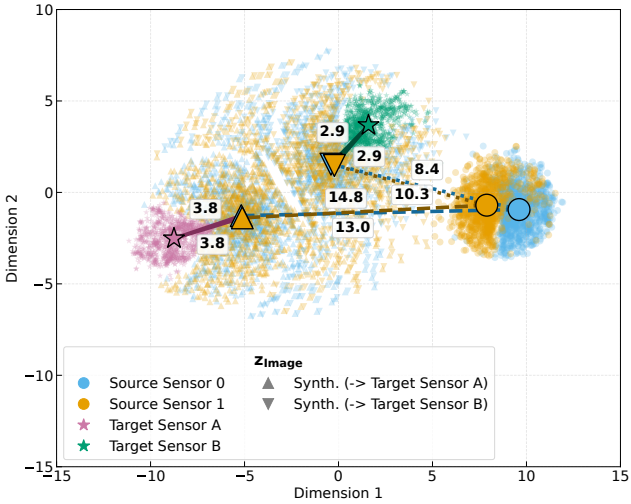}
\caption{Latent Space Distribution ($Z_{Image}$): Target Alignment via Disentanglement.}
\label{fig:tsne_z_image}
\end{figure}

We further examined the image latent space ($Z_{Image}$) to assess the effectiveness of our arithmetic-based style transfer. As shown in Figure~\ref{fig:tsne_z_image}, the distribution of the original source images is clearly separated from the target domain due to differences in lighting, gel texture, and background noise.

Upon injecting the target background vector ($Z_{Base}^{Target}$) into the source deformation representations $Z_{Deform}^{Source}$, the resulting synthetic data shifts significantly towards the real target cluster. Notably, the synthetic distribution aligns with the target domain's manifold but does not collapse entirely onto the real target data points. This separation mirrors the structure already observed in $Z_{Deform}$ (Figure~\ref{fig:tsne_z_deform}), confirming that the geometric distinction between the source indentations and the target indentations persists in the image space. This result provides strong evidence that the framework successfully transfers the target sensor's optical properties (style) while strictly preserving the original contact geometry (content).

\begin{figure}[!htb]
  \centering
    \includegraphics[width=1\columnwidth]{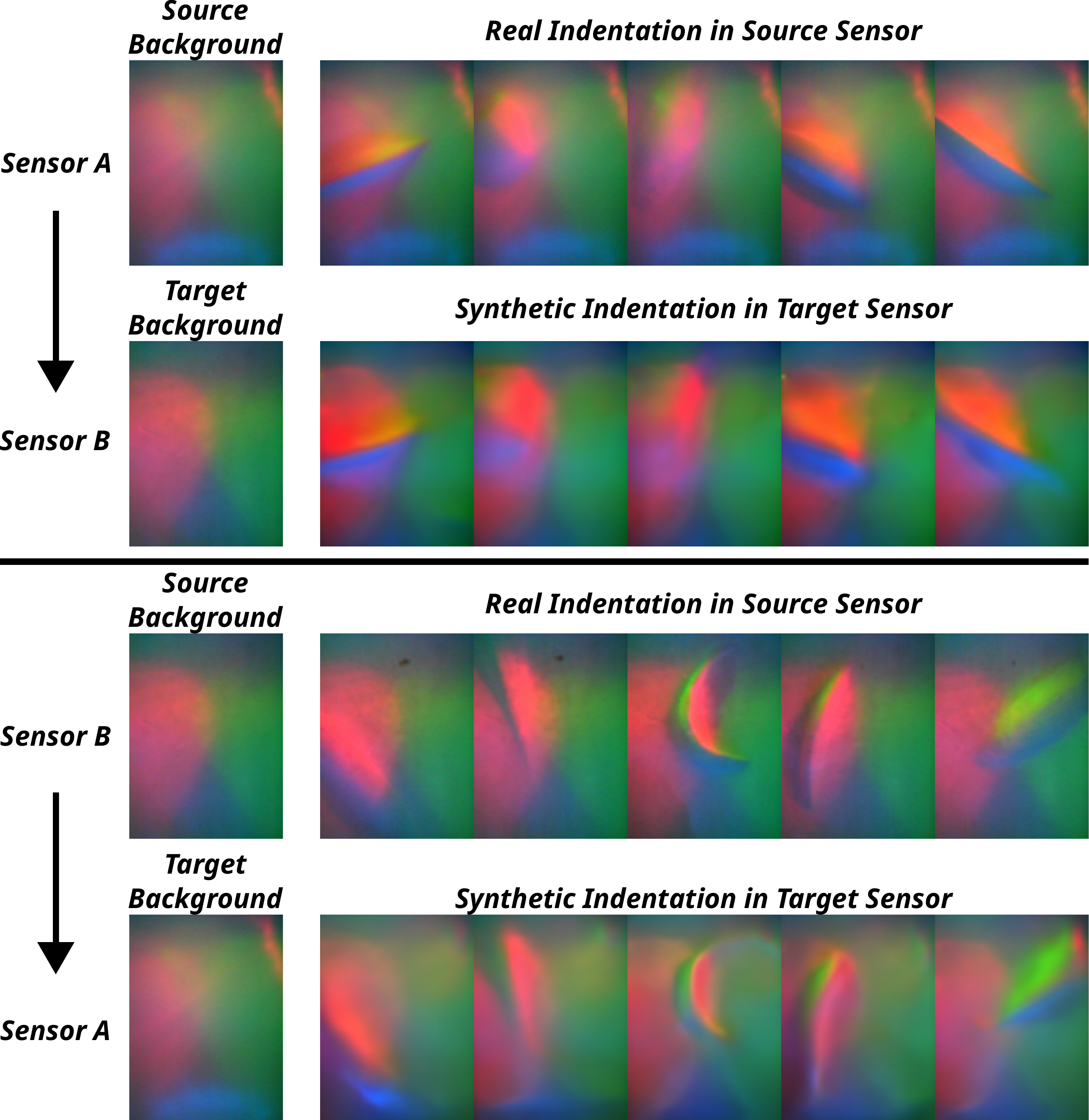}
    \caption{Qualitative generalization to fully unseen indentations and backgrounds (Sensor A $\leftrightarrow$ Sensor B).}
    \label{fig:qualitative_results}
\end{figure}

\textbf{Qualitative Generalization to Unseen Domains:} Finally, we assessed the framework's performance in the most challenging inference scenario: transferring unseen geometries from an unseen source sensor to an unseen target sensor. Because no paired ground-truth data exists for these specific zero-shot domain transfers, traditional pixel-wise metrics cannot be computed. Therefore, the evaluation of these results relies on qualitative visual inspection. We specifically assess whether the semantic contact geometry, such as the shape and edges of the indentation, is successfully preserved while the optical properties, including lighting, noise, and background, correctly match the target sensor. As illustrated in Figure~\ref{fig:qualitative_results}, the model successfully synthesizes realistic tactile images even when neither the contact shape nor the sensor background was present in the training set. Visual inspection confirms that geometric details, such as object edges, positions and orientations, are preserved from the source domain, while the global optical properties (lighting gradients, manufacturing artifacts) are rendered to closely resemble the target sensor's characteristics. We note that while the $\beta$-VAE architecture results in a slight smoothing of the output, the semantic content of the indentations remains distinct and recognizable. This demonstrates that the disentanglement strategy generalizes to fully in-the-wild scenarios.

The quantitative and qualitative results collectively demonstrate that explicit latent space arithmetic is a viable mechanism for zero-shot domain adaptation within the DIGIT sensor family. The ability to decouple $Z_{Deform}$ allows us to treat contact geometry as a modular component that can be rendered with arbitrary optical properties. This capability has profound implications for data efficiency: it enables a powerful data augmentation strategy where dataset diversity is synthetically expanded, allowing downstream models to generalize across sensor units without the prohibitive cost of collecting new physical data for every hardware variation.


\subsection{Cross-Sensor Generalization: Adapting to GelSight R1.5}

\begin{figure}[!htb]
  \centering
    \includegraphics[width=1\columnwidth]{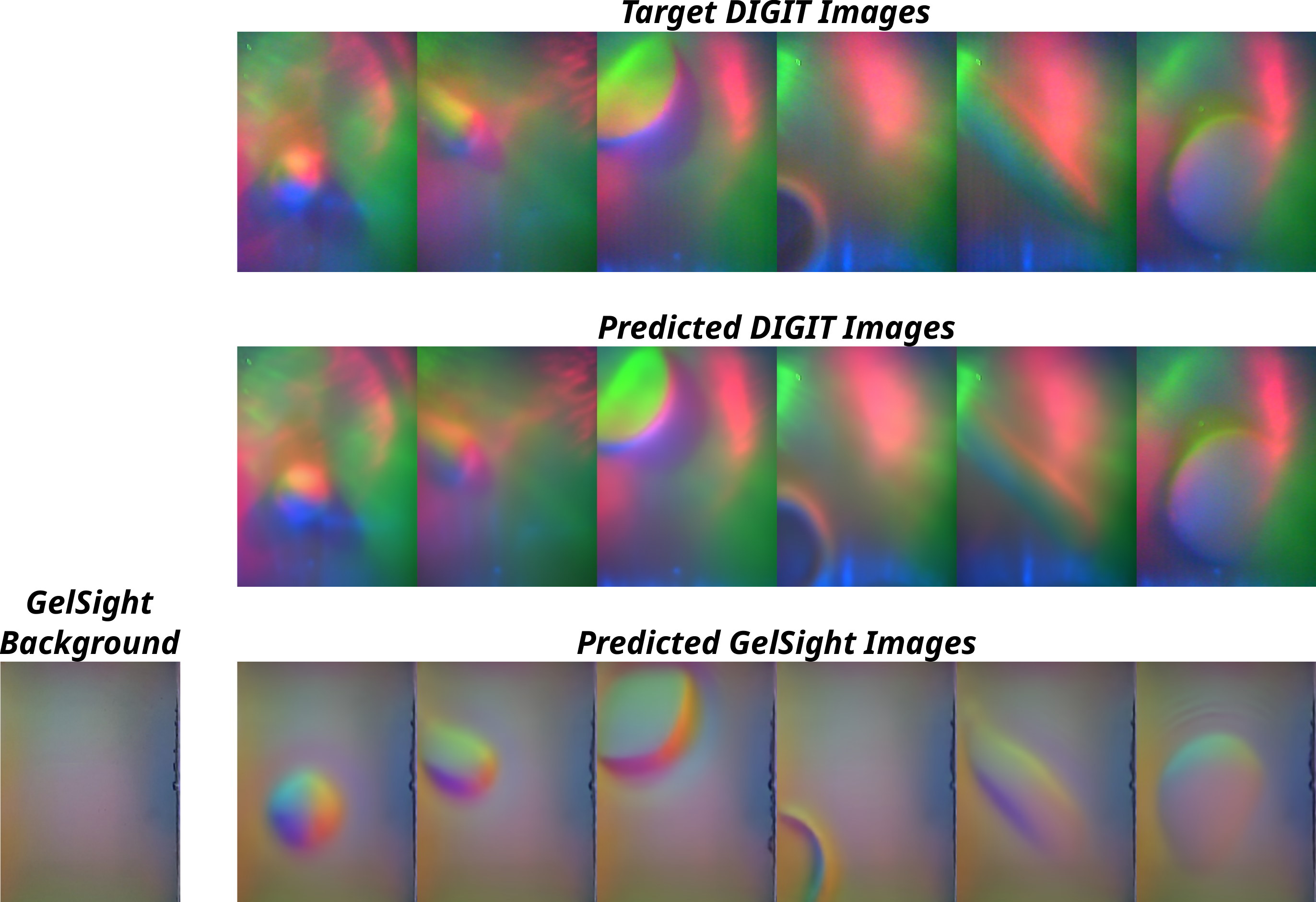}
    \caption{First row: real DIGIT images. Second row: predicted DIGIT images. Third row: predicted images using a GelSight R1.5 sensor's background image (first image left).}
    \label{fig:results_gelsight}
\end{figure}

To investigate the potential for cross-sensor generalization, as detailed in Section~\ref{sec:methods_cross_sensor}, we applied the full SPLIT pipeline to the GelSight R1.5 sensor as a preliminary proof-of-concept.
Similar to the qualitative evaluation in Section~\ref{sec:results_intra_sensor}, the absence of paired ground-truth data for a DIGIT indentation transferred onto a GelSight background means traditional metrics are inapplicable. We therefore visually assess whether the R1.5's characteristic optical properties are successfully synthesized without distorting the underlying semantic contact shape.
Figure~\ref{fig:results_gelsight} illustrates these initial results.

These results highlight the ability of our method to learn invariant latent spaces, generalizing not only across unseen DIGIT sensors but also to distinct hardware configurations like the GelSight R1.5. By freezing the geometric encoder and projection MLP, we demonstrate that the geometric deformation vector ($Z_{Deform}$) serves as a robust, sensor-agnostic representation of contact. Furthermore, this disentanglement allows for the flexible injection of arbitrary background patterns ($Z_{Base}$), enabling the synthesis of contact deformations onto diverse sensor backgrounds or noise profiles without altering the underlying geometry.

However, we note that this transferability is currently bounded by the topological similarity of the sensors. The successful transfer between DIGIT and GelSight R1.5 relies on their shared planar geometry. Significantly different form factors (e.g., the curved surface of a DIGIT 360~\citep{lambeta2024digit360}) would likely require re-aligning the geometric latent space. Additionally, while the texture synthesis is robust, certain domain-specific artifacts, such as surface ripples, are observed (e.g., in the rightmost image). A detailed discussion of these artifacts is provided in the supplementary material on our website.

The fine-tuned decoder successfully translates this geometric deformations into the target domain, accurately synthesizing the GelSight’s characteristic optical response. Notably, the model adapts to the specific lighting gradients and marker patterns of the R1.5 without distorting the underlying contact shape derived from the DIGIT simulation. This confirms that our modular architecture effectively disentangles the physical contact geometry from the sensor-specific optical noise and color profiles. The successful synthesis suggests that synthetic data generated for one sensor can be potentially repurposed for another with minimal real-world data requirements, largely bypassing the need for extensive physical recollection for every new hardware iteration.


\subsection{Computational Efficiency and Real-Time Feasibility}

Table~\ref{tab:inference_time} presents a comparative analysis of the inference latency. We report the overall time as the summation of the physics simulation step, required to generate the deformation mesh, and the subsequent image rendering step (via Pyrender~\textsuperscript{\ref{Pyrender}} for the baselines or our neural networks for SPLIT).

\begin{table*}[!tbp]
\caption{Comparison of inference times for soft-body simulation and rendering using Isaac Gym~\citep{Makoviychuk2021Isaac}, Taxim~\citep{Si2022Taxim}, FOTS~\citep{Zhao2024FOTS} and SPLIT on CPU in seconds (s).}
\label{tab:inference_time}
\centering
\renewcommand{\arraystretch}{1.0}
{
\begin{tabular}{@{}l|Y{25mm}|Y{25mm}|Y{25mm}|Y{25mm}}

\toprule
\multirow{2}{*}{\textbf{Time for}} & \textbf{Isaac Gym~\citep{Makoviychuk2021Isaac}} & \textbf{Taxim~\citep{Si2022Taxim}} & \textbf{FOTS~\citep{Zhao2024FOTS}} & \textbf{SPLIT} \\ \cmidrule{2-5}
& Mesh Simulation & \multicolumn{3}{@{}c@{}}{Mesh to Image Simulation} \\ \midrule\hline
\textbf{Low Resolution} & 0.060 &    &    & \textbf{0.066} \\ \hline
\textbf{High Resolution} &  0.870 &  0.304 &  0.247 &  \\ \hline  \midrule
\textbf{Overall time (s)} &   & 1.174 & 1.117 & \textbf{0.126} \\ \bottomrule
\end{tabular}
}
\end{table*}

The results highlight a significant efficiency advantage for our approach. As shown in Table~\ref{tab:inference_time}, the Taxim~\citep{Si2022Taxim} and FOTS~\citep{Zhao2024FOTS} baselines necessitate high-resolution meshes (approx.\ $80k$ vertices) to achieve acceptable visual quality. This dependency imposes a heavy computational burden during the physics simulation step ($0.870$~s), resulting in total inference times exceeding $1.1$~s per frame. In contrast, SPLIT is designed to operate robustly on low-resolution meshes (approx.\ $6k$ vertices). This reduces the physics simulation overhead to just $0.060$~s. When combined with our efficient neural rendering ($0.066$~s), SPLIT achieves a total inference time of $0.126$~s, representing an approximate 9$\times$ speedup over the state-of-the-art baselines while running on the same CPU hardware.

\begin{figure}[!htb]
  \centering
    \includegraphics[width=1\columnwidth]{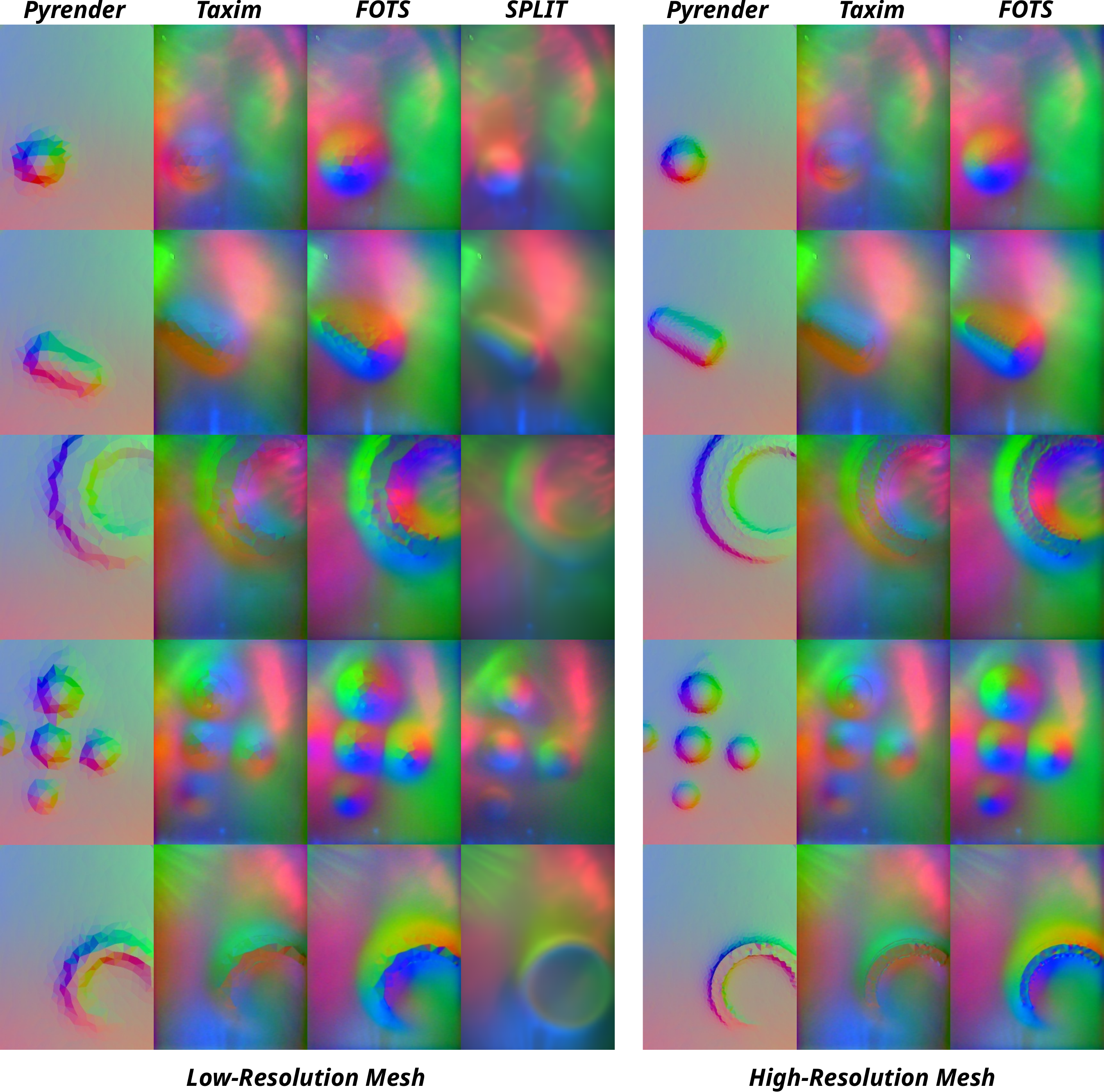}
    \caption{Comparison of simulated images generated using low-resolution ($6,103$ vertices) and high-resolution ($80,744$ vertices) mesh inputs from Pyrender~\textsuperscript{\ref{Pyrender}}, Taxim~\citep{Si2022Taxim}, FOTS~\citep{Zhao2024FOTS}, and SPLIT.}
    \label{fig:results_lowHighresolution}
\end{figure}

Figure~\ref{fig:results_lowHighresolution} illustrates the rendering quality across methods using both low- and high-resolution inputs. We observe that Taxim and FOTS produce visible discretization artifacts, manifesting as jagged edges, when the mesh density is reduced. These artifacts stem from the geometric dependence of their height-map calculation algorithms. Conversely, SPLIT effectively suppresses these structural imperfections, generating smooth, high-quality tactile images even from coarse geometric inputs. This confirms that our neural projection learns to abstract away mesh discretization errors, offering a distinct advantage for deployment in resource-constrained, real-time control environments.

\subsection{Reconstructing DIGIT Sensor Meshes}
\label{sec:Image_Mesh}

To assess the geometric fidelity of our generative pipeline, we measure the deviation between the reconstructed 3D meshes and the ground truth geometry.
Table~\ref{tab:results_mesh} presents the average results in millimeters over all samples and vertices in the test set.

\begin{table*}[!tbp]
\caption{Quantitative comparison of 3D mesh reconstruction performance between the proposed SPLIT and NOSPLIT architectures and the baseline by Zhu et al.~\citep{Zhu2022Learning}. Metrics include RMSE, $\ell_1$, and average Euclidean distance (in mm).}
\label{tab:results_mesh}
\centering
\renewcommand{\arraystretch}{1.0}
{
\begin{tabular}{@{}l|Y{27mm}|Y{27mm}|Y{27mm}@{}}
\toprule
\textbf{Methods} & \textbf{RMSE (mm) $\downarrow$} & \textbf{$\boldsymbol{\ell_1}$ (mm) $\downarrow$} &  \textbf{Euc.\ Dist.\ (mm) $\downarrow$} \\ \midrule \hline
\textbf{Zhu et al.~\citep{Zhu2022Learning}} & 0.107 & 0.047 & 0.099 \\ \hline
\textbf{NOSPLIT} & 0.111 & 0.053 & 0.114 \\ \hline
\textbf{SPLIT} & \textbf{0.075} & \textbf{0.028} & \textbf{0.061} \\ \hline

\end{tabular}
}
\end{table*}

We observe that our SPLIT framework outperforms the baseline method proposed by Zhu et al.~\citep{Zhu2022Learning} across all metrics. This performance gap highlights the advantage of training on physically simulated meshes paired with real sensor images, compared to the synthetic approximations used by the baseline. By using realistic physics and real-world visual data, our model reconstructs mesh geometry with higher precision.

Furthermore, the internal comparison validates the effectiveness of our disentanglement strategy. The SPLIT approach achieves the lowest reconstruction error (RMSE of $0.075$ mm), outperforming the NOSPLIT direct mapping (RMSE of $0.111$ mm). Crucially, both the Direct mapping and Zhu et al. exhibit significant performance degradation when applied to unseen DIGIT sensors. Because these methods lack an explicit mechanism to separate geometry from optics, they overfit to the specific optical characteristics, such as the unique lighting gradients, of the sensors used during training. In contrast, by explicitly subtracting the optical background ($Z_{Base}$) to isolate the deformation features ($Z_{Deform}$), our method removes these sensor-specific artifacts. This allows the network to focus exclusively on the deformation patterns that correlate with the 3D shape, leading to robust 3D reconstruction even on novel DIGIT units.

\begin{figure}[!htb]
  \centering
    \includegraphics[width=0.65\columnwidth]{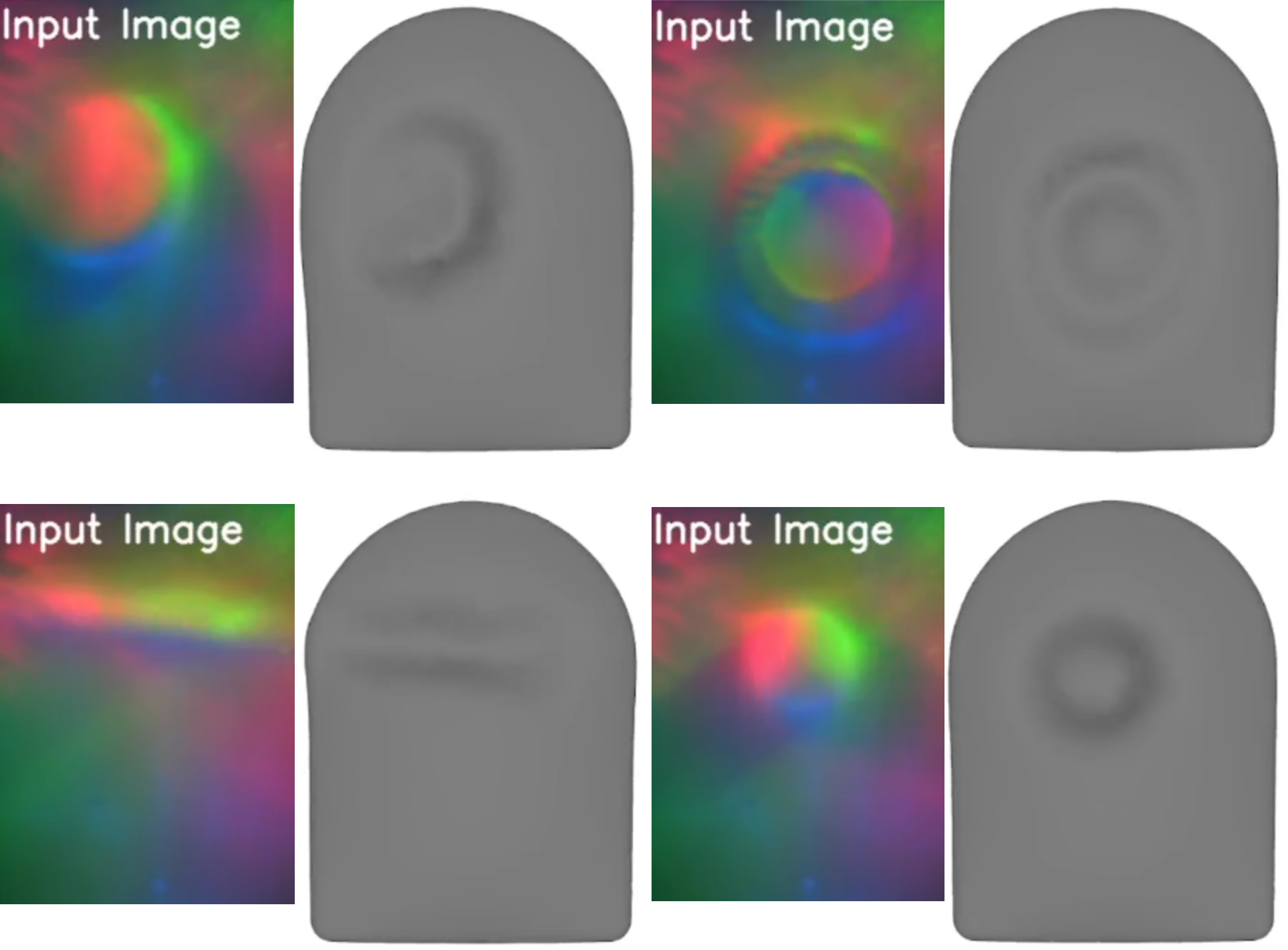}
    \caption{Qualitative comparison between real tactile input images and their corresponding 3D mesh reconstructions.}
    \label{fig:mesh_results_visualization}
\end{figure}

To further validate the physical consistency of our method under natural contact conditions, we evaluate the reconstructed mesh deformations directly from real sensor inputs. Figure~\ref{fig:mesh_results_visualization} presents qualitative visualizations comparing real tactile images with their corresponding 3D mesh reconstructions. These examples demonstrate that our framework accurately captures the topological deformations caused by varying real objects, confirming that the simulated meshes align seamlessly with actual physical interactions. Additionally, further dynamic examples can be seen in our webpage, which showcases the algorithm operating in real-time. However, it should be noted that reconstructing complex, multi-part objects or those exhibiting novel, uncommon textures remains challenging. While the model can successfully interpolate deformations for objects that closely resemble the basic indenters from our training set, it struggles to generalize to highly irregular shapes or random, unseen textures. Consequently, accurately rendering these intricate, out-of-distribution interactions remains a compelling direction for future research.

Finally, to investigate the stability of our bidirectional capabilities (Image $\rightarrow$ Mesh $\rightarrow$ Image) for potential long-term temporal state estimation tasks, we conducted a continuous cyclic reconstruction analysis. By feeding the output of one modality back as the input to the other over 40 iterations, we tracked the manifestation of cumulative errors quantitatively using both absolute deviation and cycle-to-cycle drift metrics.

\begin{figure}[!htb]
  \centering
    \includegraphics[width=1\columnwidth]{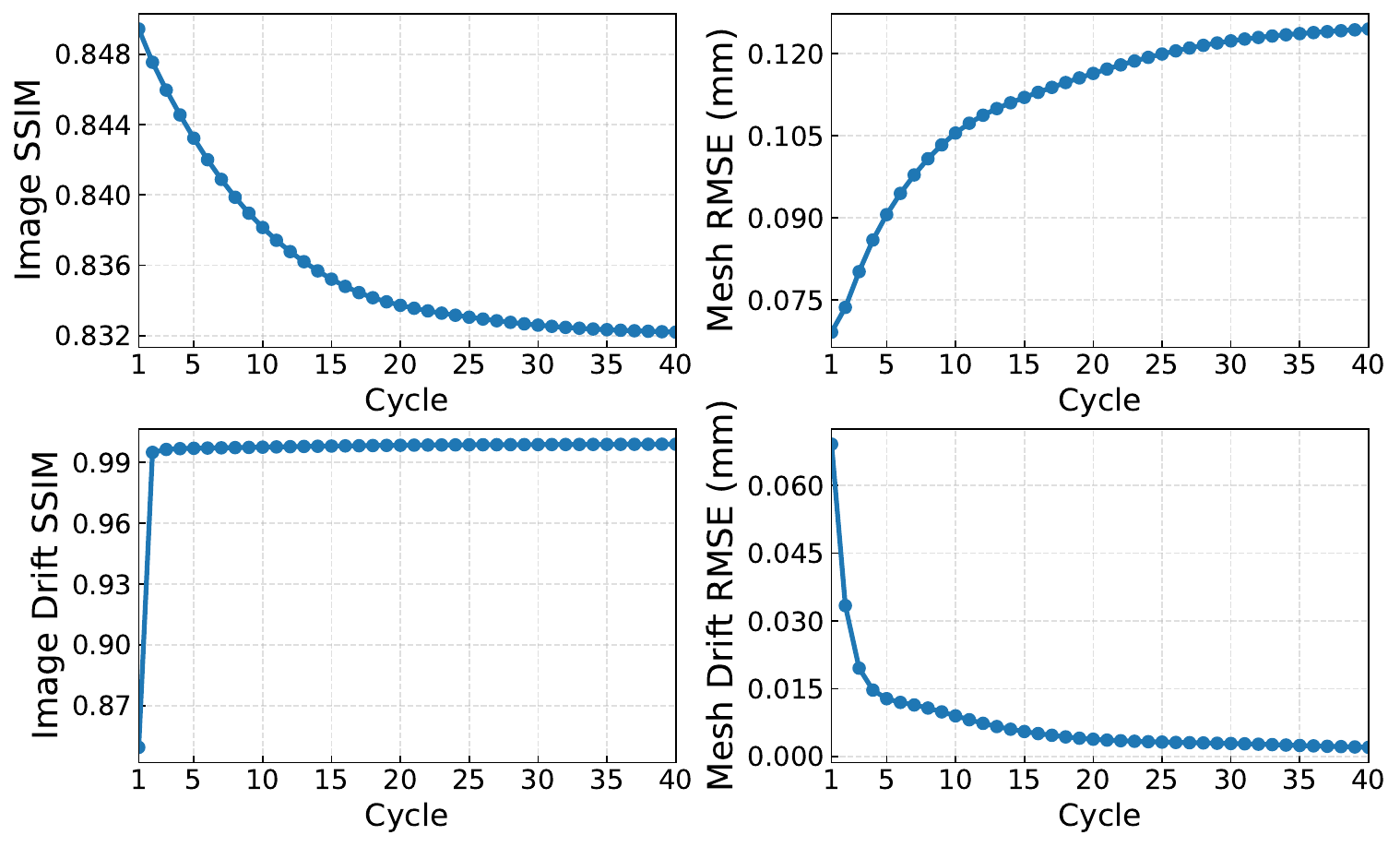}
    \caption{Quantitative evaluation of cyclic reconstruction stability over 40 iterations. The top row displays absolute metrics (Image SSIM and Mesh RMSE), which measure the total deviation of each cycle's output relative to the original real-world input. The bottom row displays drift metrics, calculating the change between consecutive cycles (Cycle $N$ vs.\ Cycle $N-1$).}
    \label{fig:cyclic_loss}
\end{figure}

\begin{figure*}[!tbp]
  \centering
    \includegraphics[width=0.8\textwidth]{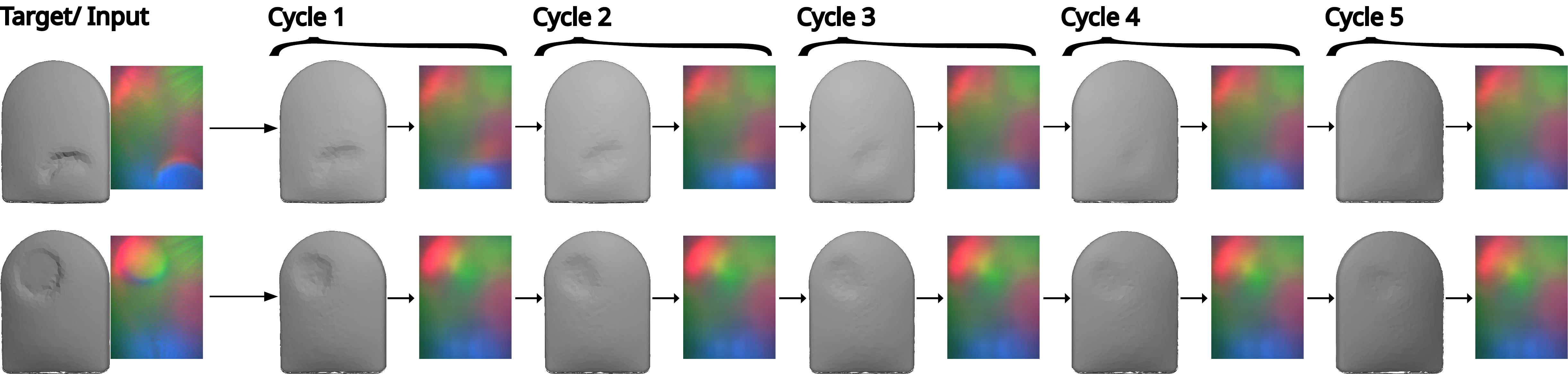}
    \caption{Visual representation of the continuous cyclic reconstruction process over five consecutive iterations. The leftmost column presents the initial target inputs, consisting of the 3D mesh and the corresponding tactile image. The subsequent columns illustrate the recursively generated meshes and tactile images for Cycle 1 through Cycle 5. The top and bottom rows display two distinct contact deformation examples.}
    \label{fig:cyclic_visualizations}
\end{figure*}

Our analysis reveals a stark contrast between the initial translation and subsequent recursive inferences. In the first cycle, when translating directly from real images to synthetic meshes (or from real meshes to synthetic images) the framework successfully captures the precise geometric shape, macroscopic position, and structural details with high fidelity. However, the performance degrades immediately when generating from recursive synthetic inputs (i.e., mapping a synthetic mesh to a synthetic image, or vice versa). As soon as the system processes its own synthetic outputs, the precise geometric shape is smoothed into a more blob-like representation, retaining only the general position. As the cycles continue, the generated images become progressively blurrier, and the reconstructed meshes gradually regress toward the undeformed state, as visualized in Figure~\ref{fig:cyclic_visualizations}.

The quantitative metrics reveal two distinct phases of this degradation, as illustrated in Figure~\ref{fig:cyclic_loss}. First, the absolute error metrics (top row), which compare each cycle's output to the original real-world input, steadily worsen over time: image structural similarity index measure (SSIM) decreases while the 3D mesh root-mean-square error (RMSE) increases. This confirms the progressive loss of high-frequency details as the visual outputs blur and the meshes regress toward the undeformed state. Second, the drift metrics (bottom row), which calculate the delta between consecutive cycles (Cycle $N$ vs.\ Cycle $N-1$), demonstrate rapid convergence. The Image Drift SSIM immediately approaches 1.0, which visually manifests as a direct transition to a blurry, blob-like image, while the Mesh Drift RMSE decays quickly toward zero. This indicates that rather than becoming numerically unstable, the network quickly converges to a stable equilibrium, locking into the smoothed representation after the initial cycles.

We attribute this compounding error to the inherent design of the $\beta$-VAE architecture, specifically its information bottleneck and stochastic sampling mechanism. The strong bottleneck required to achieve disentanglement forces the model to prioritize global features over fine details. Furthermore, because the latent representation relies on Gaussian sampling, each recursive encode-decode step injects slight structural variance. In a continuous loop, this repeated stochastic sampling acts as a cumulative spatial smoothing effect, progressively washing out sharp, deterministic edges. However, it is crucial to emphasize that these exact architectural traits are precisely what make the $\beta$-VAE ideal for our primary objective. In a single-step projection, this information bottleneck acts as an essential feature extractor, successfully stripping away sensor-specific optical noise to enable the robust zero-shot domain transfer and latent arithmetic central to our framework. Consequently, while SPLIT is highly effective for single-step inference from real data, solving this progressive smoothing for continuous temporal tracking falls outside the scope of the current paper and remains an important area for future development.

\subsection{Downstream Task: Force Estimation}
\label{sec:force_prediction}

To validate the efficacy of SPLIT for downstream robotic applications, we quantify the physical fidelity of its representations by analyzing force estimation accuracy across different input modalities. Specifically, we investigate whether projecting tactile images into a disentangled geometric latent space yields more robust force predictions than relying on raw visual data, particularly when generalizing to unseen sensors.
Table~\ref{tab:force_prediction} summarizes the mean MAE values and standard deviations of the predicted force vectors, measured in Newtons.

\begin{table*}[tbp]
\caption{Force prediction results: Mean Absolute Error (MAE) values in Newtons (N) for different network inputs, including mesh vertices, image, mesh latent space vector, and image latent space vector with and without vector arithmetic (SPLIT and NOSPLIT). The results are calculated over 20 different seeds. Lower MAE values indicate better performance. The values in parentheses represent the standard deviation.}
\label{tab:force_prediction}
\centering
\renewcommand{\arraystretch}{1.0}
{
\begin{tabular}{@{}p{45mm}|Y{23mm}|Y{23mm}|Y{23mm}|Y{23mm}@{}}
\toprule
\textbf{Networks Input} & \textbf{$F_x$ (N) $\downarrow$} & \textbf{$F_y$ (N) $\downarrow$} & \textbf{$F_z$ (N) $\downarrow$} & \textbf{$||F||_2$ (N) $\downarrow$}   \\ \midrule
\hline
\textbf{Raw Vertices} & \textbf{0.257 (0.007)} & \textbf{0.288 (0.007)} & 1.360 (0.203) & 1.369 (0.205) \\ \hline
\textbf{Raw Images} & 1.049 (0.161) & 1.090 (0.191) & 6.046 (2.043) & 6.092 (2.045) \\ \hline

\textbf{Mesh Latent} & 0.273 (0.004) & 0.308 (0.003) & \textbf{1.125 (0.017)} & \textbf{1.130 (0.018)}\\ \hline
\textbf{Image Latent (NOSPLIT)} & 0.536 (0.055) & 0.611 (0.075) & 2.881 (0.259) & 2.875 (0.254) \\ \hline
\textbf{Image Latent (SPLIT)} & 0.386 (0.005) & 0.462 (0.002)  & 1.620 (0.042) & 1.637 (0.043)\\ \hline
\textbf{Projected Mesh (NOSPLIT)} & 0.578 (0.033)  & 0.771 (0.054)  & 2.341 (0.048)  & 2.357 (0.049) \\ \hline
\textbf{Projected Mesh (SPLIT)} & 0.391 (0.002)& 0.473 (0.003)& 1.632 (0.029)& 1.641 (0.030) \\ \hline
\end{tabular}
}
\end{table*}

We first observe a substantial performance gap between the raw baselines. The Raw Vertices model achieves a relatively low MAE ($1.369$ N), whereas the Raw Images model suffers significant degradation, yielding the highest error of $6.092$ N. This difference arises because the test set contains held-out DIGIT sensors with unique optical characteristics (e.g., lighting gradients, color shifts) not seen during training. While 3D meshes are naturally invariant to these optical shifts, raw images contain sensor-specific noise that distracts the network, leading to poor generalization.

In contrast, our results demonstrate that learned latent representations effectively distill force-relevant features, offering a distinct advantage over raw image inputs. Using the fixed latent embeddings of our VAEs as input leads to a sharp reduction in error, with the Mesh Latent ($Z_{Mesh}$) achieving the lowest overall MAE ($1.130$ N). Critically, our result confirms the value of our explicit disentanglement strategy. When predicting force from image embeddings, the arithmetic latent embedding (SPLIT) ($1.637$ N) significantly outperforms the direct embedding (NOSPLIT) ($2.875$ N). This indicates that the holistic visual state ($Z_{Image}$) contains entangled optical information, such as background lighting, that degrades force estimation. By using latent arithmetic to subtract the background and isolate the deformation signal ($Z_{Deform}$), we allow the network to focus purely on the physical interaction, resulting in greater robustness to sensor variations.

Building on this, our full pipeline, projecting images into the mesh latent space, successfully preserves physical fidelity even when applied to unseen sensors. The Projected Mesh (SPLIT) mapping achieves a total MAE of $1.641$~N, significantly outperforming the direct mapping (NOSPLIT) ($2.357$~N). Importantly, this result is statistically equivalent to the performance of the Image Latent (SPLIT) ($1.637$~N). This minimal gap indicates that the projection network successfully translates visual data into the geometry-centric domain with negligible loss of critical physical information. Overall, these findings demonstrate that SPLIT’s vector arithmetic strategy effectively filters out hardware-specific optical noise, enabling accurate estimation of forces on unseen sensors where raw image methods typically fail.

Beyond the total force magnitude ($||F||_2$), we analyze specific force components to validate the model's sensitivity. We observe that the Raw Vertices baseline achieves the absolute lowest error for shear forces ($F_x = 0.257$ N, $F_y = 0.288$ N), confirming that direct access to 3D mesh geometry provides the highest fidelity for detecting subtle lateral deformations. However, the Mesh Latent model outperforms the Raw Vertices on the dominant normal component ($F_z = 1.125$ N vs. $1.360$ N). We attribute this improvement to the VAE's ability to encode the global indentation shape into a compact representation, effectively smoothing out local irregularities that can distort the lateral force estimates derived from raw geometry.
Despite this trade-off, the Mesh Latent representation remains highly precise across all axes. Its normal force error ($1.125$ N) represents approximately 9.4\% of the sensor's operating range ($1$–$13$ N). Meanwhile, its shear force error ($0.273$ N for $F_x$) corresponds to a relative error of just 4.6\% (range $\pm 3$ N), closely trailing the Raw Vertices baseline. This confirms that the model successfully balances robust normal force estimation with the high precision necessary for slip detection and delicate manipulation tasks.


\section{Limitations and Future Work}
While promising, our current approach has limitations. First, the framework's geometric accuracy is bounded by the fidelity of the underlying Isaac Gym soft-body simulation and the specific properties of the calibrated DIGIT gel. Although calibrated, the simulation may not capture complex non-linear material behaviors or complex surface friction present in the real world. Future works could incorporate alternative physics engines, such as MuJoCo~\citep{todorov2012mujoco}, to augment the training data with more diverse physical interactions.
Second, while the image generation generalizes well across optical domains, the method is currently bounded by geometric similarity. The framework assumes comparable contact topologies (e.g., planar surfaces). Consequently, transfer to sensors with distinct form factors, such as curved or cylindrical gels, or differing sizes, is not directly supported and requires further investigation into geometric re-mapping strategies. We plan to extend this work by developing a fully sensor-agnostic mesh latent space representation to bridge these topological gaps.
Third, generalizing to highly intricate, out-of-distribution physical interactions remains an ongoing challenge. Because our framework was trained predominantly on a dataset of basic indenters, the model successfully interpolates deformations for objects that closely resemble this training set, but struggles to reconstruct complex, multi-part objects or those exhibiting entirely novel textures. Expanding the training distribution to accurately capture these complex interactions represents an important next step for this framework.

Additionally, an inherent limitation of the current bidirectional framework arises during continuous, cyclic reconstruction. As detailed in Section~\ref{sec:Image_Mesh}, while the system achieves high geometric and optical fidelity during the initial translation from real inputs, it struggles with recursive synthetic-to-synthetic inferences. The strong information bottleneck imposed by the $\beta$-VAE architecture prioritizes global positioning over fine geometric details. In a cyclic loop, feeding synthetic outputs back into the network acts as a repeated spatial smoothing operation. This causes a rapid loss of precise shape as soon as recursive generation begins, leading to a compounding signal degradation where images blur and meshes regress toward the undeformed state. While extending the framework for long-term sequential tracking is beyond the primary focus of this paper, future work will explore adjusted architectures or temporal consistency losses to mitigate this fading effect and better preserve fine structural details over sequential inferences.

Looking ahead, we aim to leverage SPLIT for rigorous Sim-to-Real transfer in closed-loop robotic manipulation. Given the high physical fidelity observed in our force estimation results, we believe our generated data can serve as a potent training source for policy learning, reducing the reliance on costly real-world data collection. Ultimately, SPLIT offers a scalable, efficient pathway toward generalizable tactile perception, opening new opportunities for adaptive robotic interaction in unstructured environments.

\section{Conclusion}
\label{sec:conclusion}
We presented SPLIT, a novel approach for simulating image-based tactile sensors. By leveraging latent space vector arithmetic to explicitly disentangle geometry from optics, our method bridges the gap between physical deformation and sensor-specific response. Our extensive evaluation demonstrates that SPLIT outperforms state-of-the-art baselines in simulation fidelity and inference speed, achieving a $9\times$ speedup while maintaining robustness to low-resolution geometric inputs. Crucially, this explicit separation of contact geometry from optical properties enables the seamless synthesis of diverse sensor outputs, including the GelSight R1.5, from a single deformation source, thereby eliminating the need for tedious manual calibration.

Beyond simulation, we validated the framework's utility in a downstream task. Our experiments in 3D mesh reconstruction and force estimation confirmed that projecting visual data into a disentangled, geometry-centric latent space significantly improves robustness to domain shifts. By filtering out optical noise, SPLIT enables accurate inference on unseen sensors where traditional raw-input methods fail. To support further research, we have released our comprehensive dataset, the calibrated soft-body Isaac Gym simulation environment, and the full codebase.


\section*{Declarations}
\subsection*{Acknowledgment} This research was partially funded by the Bundesministerium Forschung, Technologie und Raumfahrt (BMFTR) under the Programme \href{https://www.bmbf.de/DE/Forschung/TransferInDiePraxis/DeutscheAgenturFuerTransferUndInnovation/Datipilot/datipilot_node.html}{DATIPilot} Innovationssprints project No.\ 03DPS1242A (\href{https://nicolas-navarro-guerrero.github.io/projects/vibrosense/}{Vibro-Sense}).
We also want to thank AI Research at Meta and GelSight Inc.\ for the awarded DIGIT sensors via the DIGIT CFP 2024.

\subsection*{Conflict of interest/Competing interests} The authors declare that they have no conflict of interest.

\printcredits

\bibliographystyle{cas-model2-names}

\bibliography{references}

\clearpage

\begin{appendices}

\setcounter{section}{0}   
\renewcommand{\thesection}{\Alph{section}}

\section{Networks Architectures}
\label{sec:Networks_Architectures}

\subsection{Mesh Network}
Our 3D mesh disentangled variational autoencoder ($\beta$-VAE) uses a graph Convolutional autoencoder (CoMA) architecture, which leverages Fourier transformations to extract spectral information from the mesh graph. This information is then processed using Chebyshev filters, enabling localized convolution operations~\citep{Defferrard2016Convolutional}.
The encoder module consists of four graph convolutional layers, each with filter sizes of [16, 16, 16, 32] and a kernel size of 6. Downsampling operations with a factor of 2 progressively reduce the mesh's spatial dimensions. We use \textit{ReLU} activation functions. The latent space is parameterized by two parallel linear layers, each with 128 units. They compute the mean ($\mu$) and log variance $(log(\sigma^{2}))$ of a normal distribution. The decoder mirrors the encoder, but replaces downsampling with upsampling. This symmetric enables efficient mesh reconstruction. The network is trained using the Adam optimization algorithm with an initial learning rate of $0.001$, which decays linearly with a factor of 0.99, and a batch size of 128 meshes is used during training. To balance reconstruction accuracy and latent space complexity, the KL divergence term is weighted by a factor of $0.005$.
We minimize the following loss function:
\begin{equation}\label{eq:loss_mesh_vae}
    \ell_M= \textit{MSE} (M - \hat{M})+\beta_M \operatorname{KL}\left(f\left(Z_{Mesh} \mid M\right) \| \mathcal{N}(0,1)\right),
\end{equation} 
where $f$ is the encoder of the mesh $\beta$-VAE, $\beta_M$ is the weight of the KL divergence loss to control the influence of that term on the overall loss. $Z_{Mesh}$ is the sampled latent vector given the input mesh $M_i$. The mean-squared error (MSE) between the original and reconstructed meshes is calculated, considering all 3D vertices and batch samples. We implement early stopping and restrict training to 300 epochs to mitigate overfitting.

\subsection{Image Network}
Our image $\beta$-VAE model employs a convolutional neural network (CNN) architecture, comprising ResNet blocks, to learn a robust and compact representation of input images. The normalized input images have dimensions of $320 \times240\times3$. The encoder consists of multiple ResNet blocks with varying layer configurations: $[3, 3, 5, 5, 3, 3]$, each followed by downsampling layers with factors of 2, 2, 2, 2, and 5. This hierarchical design allows the model to capture a wide range of features.
The decoder mirrors the encoder's architecture, replacing downsampling with upsampling to reconstruct the original image. Throughout the network, we employ the \textit{tanh} activation function.
For hyperparameter settings, we annealed $\beta$ over 50 epochs to a final value of $0.001$. The latent dimension $Z_{Image}$ is fixed at $256$, allowing for a compact representation of the input images. We initialize the learning rate to $0.001$ and apply linear decay with a factor of $0.99$, and weight decay with factor $0.00001$, which help stabilize training and prevent overfitting.
We train our model for up to 300 epochs, implementing early stopping to monitor performance on the validation set and prevent overfitting. This setup enables our image $\beta$-VAE model to effectively learn a compact representation of the input images while preserving their essential features.
We minimize the following loss function:
\begin{equation}\label{eq:loss_image_vae}
    \ell_I= \textit{MSE} (I - \hat{I})+\beta_I \operatorname{KL}\left(g\left(Z_{Image} \mid I\right) \| \mathcal{N}(0,1)\right),
\end{equation} 
where $g$ is the encoder of the image $\beta$-VAE and $\beta_I$ is the weight of the KL divergence loss.

\subsection{Projection Network}
We train two distinct latent space projection networks: one for projecting meshes' latent space vectors to images' latent space vectors (Mesh-to-Image) and one for the reverse (Image-to-Mesh). Both networks are implemented as Multilayer Perceptrons (MLPs), consisting of 4 layers. The number of neurons in each layer is as follows: [512, 1024, 1024, 256], utilizing the \textit{ELU} activation function. 
Dropout regularization is applied after the first two layers, with dropout rates of 0.2 and 0.4,  to prevent overfitting. This strategy helps the networks to learn robust representations of the input data. Both networks use a batch size of 512 and minimize the MSE between the predicted and the target values. The architecture of our latent space projection networks is designed to effectively capture the complex relationships between meshes and images, allowing for accurate projections and reconstructions.

\subsection{Force Prediction Network}
We utilize a unified architecture for all force estimation tasks presented in this paper. The network is implemented as a four-layer MLP with layer sizes of [128, 512, 512, 128]. We employ the \textit{ELU} activation function across all layers. To prevent overfitting and encourage robust representation learning, dropout regularization is applied after the first two layers with a rate of 0.2. The network is trained using a batch size of 512 and a learning rate of $0.001$, minimizing the MSE between the predicted and ground-truth force vectors.

\section{Analysis of Visual Artifacts in GelSight Reconstruction}
\label{sec:appendix_ripples}

\begin{figure}[!htbp]
  \centering
    \includegraphics[width=1\columnwidth]{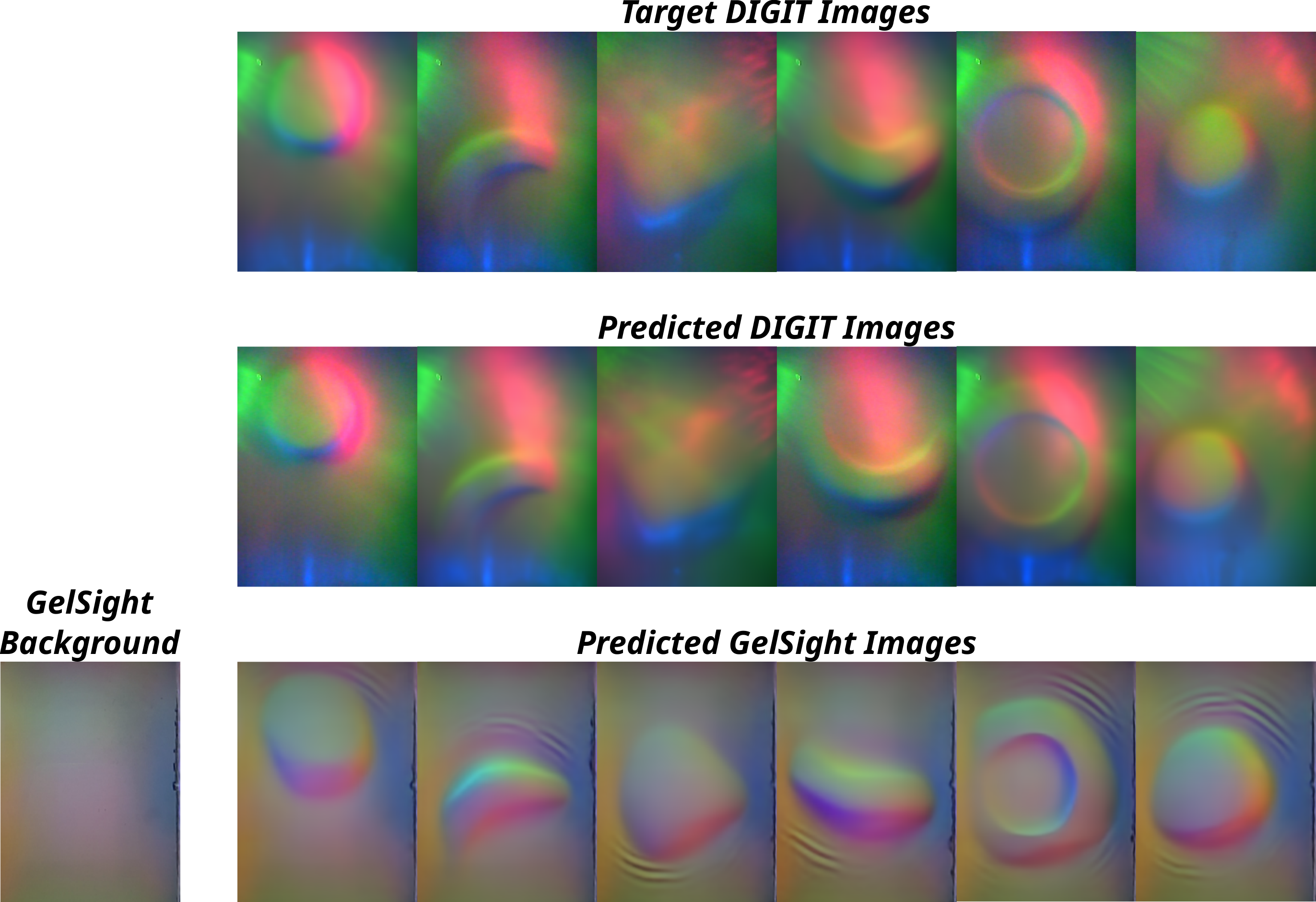}
    \caption{Visual analysis of domain-specific artifacts. First row: Ground-truth images in the DIGIT domain. Second row: Predicted contact images in the DIGIT domain. Bottom row: Cross-sensor predictions projected into the GelSight R1.5 domain, showing characteristic texture ripple patterns on the surface.}    
    \label{fig:results_gelsight_ripples}
\end{figure}

While the cross-sensor generalization to GelSight R1.5 demonstrates strong structural alignment, closer inspection of some of the generated samples reveals artifacts resembling surface ripples (see Figure~\ref{fig:results_gelsight_ripples}, third row). In real-world GelSight imagery, such ripples are typically caused by lateral shear forces deforming the elastomeric coating~\citep{Lengiewicz2020Finite}. However, in our generated results, the orientation and magnitude of these ripples do not strictly align with the applied shear forces. We attribute this discrepancy to the distribution of the training data used for fine-tuning. The \textit{ObjectFolder-Real} dataset~\citep{Gao2023Objectfolder} contains recordings rich in shear-induced deformations. Since our fine-tuning process relies on a $\beta$-VAE to adapt the decoder to this visual domain, the model learns to reconstruct these textures as intrinsic features of the GelSight style. Consequently, the network hallucinates these ripple patterns to match the statistical distribution of the target domain, rather than explicitly deriving them from the input physics. While this enhances the photorealism of the texture, the ripples should be interpreted as a learned dataset bias rather than a physically accurate simulation of shear dynamics in this specific transfer setting.

\end{appendices}

\end{document}